\documentclass[10pt,twocolumn,letterpaper]{article}

\usepackage[pagenumbers]{iccv} 

\usepackage{bm}
\usepackage{bbm}
\usepackage{amsmath}
\usepackage{mathtools}
\usepackage{amssymb}
\usepackage[linesnumbered,ruled]{algorithm2e}
\usepackage{algorithmic}
\usepackage{dirtytalk}
\usepackage{gensymb}

\definecolor{iccvblue}{rgb}{0.21,0.49,0.74}
\PassOptionsToPackage{pagebackref,breaklinks,colorlinks,allcolors=iccvblue}{hyperref}\usepackage{hyperref}

\def\paperID{*****} 
\def\confName{ICCV}
\def\confYear{2025}

\title{Interpretable deformable image registration:  A geometric deep learning perspective}

\author{%
  Vasiliki Sideri-Lampretsa\(^*\)~\orcidlink{0000-0003-0135-7442} \\
  Technical University Munich\\
  Munich, Germany \\
  \texttt{vasiliki.sideri-lampretsa@tum.de} \\ 
  \and
  Nil Stolt-Ansó\(^*\)~\orcidlink{0009-0001-4457-0967} \\
  Technical University Munich\\
  Munich, Germany \\
  \texttt{nil.stolt@tum.de} \\
  \and
Huaqi Qiu~\orcidlink{0000-0002-9706-7310} \\
  Technical University Munich\\
  Munich, Germany \\
  \texttt{harvey.qiu@tum.de} \\
  \and
  Julian McGinnis~\orcidlink{0009-0000-2224-7600} \\
  Technical University Munich\\
  Munich, Germany \\
  \texttt{julian.mcginnis@tum.de} \\
  \and
  Wenke Karbole \\
  Technical University Munich\\
  Munich, Germany \\
  \texttt{wenke.karbole@tum.de} \\
  \and
  Martin Menten~\orcidlink{0000-0001-8261-7810} \\
  Technical University Munich\\
  Munich, Germany \\
  \texttt{martin.menten@tum.de} \\
  \and
  Daniel Rueckert~\orcidlink{0000-0002-5683-5889} \\
  Technical University Munich\\
  Munich, Germany \\
  \texttt{daniel.rueckert@tum.de} \\
}

\begin{document}
\maketitle
\begin{abstract}
Deformable image registration poses a challenging problem where, unlike most deep learning tasks, a complex relationship between multiple coordinate systems has to be considered.
Although data-driven methods have shown promising capabilities to model complex non-linear transformations, existing works employ standard deep learning architectures assuming they are general black-box solvers.
We argue that understanding how learned operations perform pattern-matching between the features in the source and target domains is the key to building robust, data-efficient, and interpretable architectures.
We present a theoretical foundation for designing an interpretable registration framework: separated feature extraction and deformation modeling, dynamic receptive fields, and a data-driven deformation functions awareness of the relationship between both spatial domains.
Based on this foundation, we formulate an end-to-end process that refines transformations in a coarse-to-fine fashion.
Our architecture employs spatially continuous deformation modeling functions that use geometric deep-learning principles, therefore avoiding the problematic approach of resampling to a regular grid between successive refinements of the transformation.
We perform a qualitative investigation to highlight interesting interpretability properties of our architecture.
We conclude by showing significant improvement in performance metrics over state-of-the-art approaches for both mono- and multi-modal inter-subject brain registration, as well as the challenging task of longitudinal retinal intra-subject registration.
We make our code publicly available\footnote{\href{https://anonymous.4open.science/status/GeoReg-1A1D}{https://anonymous.4open.science/status/GeoReg-1A1D}}.

\end{abstract}    
\section{Introduction}
\label{sec:intro}

Image registration is an indispensable tool in medical image analysis that aligns anatomically or functionally corresponding regions across images, often from different modalities and time points~\citep{Sotiras2013DeformableMI}. 
In particular, deformable registration aims to estimate a non-linear transformation that maps the \textit{source} image to the coordinate system of the \textit{target} image.

\begin{figure*}
  \centering
  \includegraphics[width=1.\linewidth]{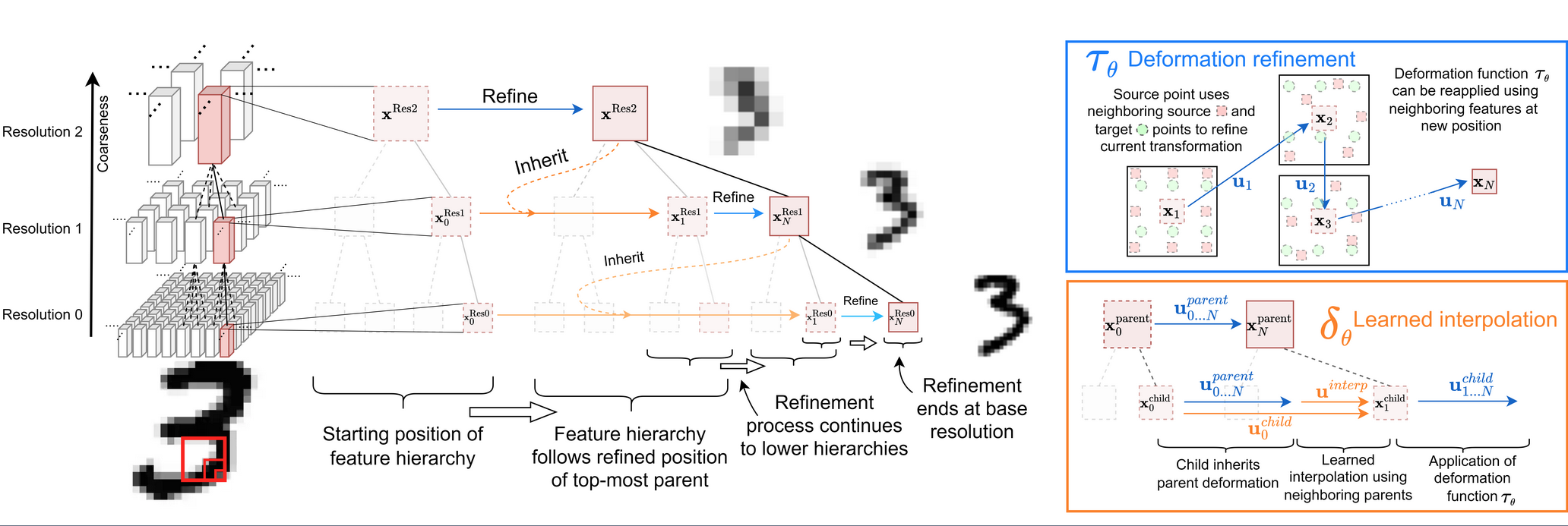}
  \caption{Our multi-resolution architecture begins by extracting features at increasingly coarse resolutions. In a coarse-to-fine fashion, the deformation function \(\tau_\theta\) refines the predicted deformation over N steps within the current resolution while the learned interpolation function \(\delta_\theta\) carries deformations onto the subsequent resolution. The architecture is supervised such that the majority of the deformation is modeled at the coarser (earliest) resolutions. Supervision of the finest (latest) resolutions provides learning signal to all deformations at the coarser levels.}
  \label{fig:overview}
\end{figure*} 

Since the advent of deep learning (DL), data-driven methods have been proposed~\citep{haskins2020,xiao2021review} to leverage learned transformation priors over an image cohort, reducing the search space of plausible transformations.
DL approaches excel at creating highly expressive, task-specific feature-extracting processes using end-to-end supervision signals.
The underlying workhorse of these architectures, the \textit{convolution}, is a pattern-matching tool that has been heavily optimized explicitly for grid-based inputs such as images. 
These convolutional layers are stacked sequentially to give rise to hierarchies of features, whereby early layers detect the presence of simple spatial building blocks (such as edges and corners), while later layers use these basic structures to capture larger-spanning spatial features of higher complexity. 

Unlike tasks such as image segmentation, where the feature extraction process is constrained to a single coordinate system, image registration poses an interesting challenge in deep learning due to having to simultaneously consider spatial relationships of multiple input coordinate systems.

\subsection{Disentangling feature extraction and deformation modeling}

Many data-driven works have adopted a \textit{single-stream} approach for deformation modeling, whereby source and target images are simply concatenated in the channel dimension as input to a convolutional network~\citep{balakrishnan2019,Dalca2018UnsupervisedLF,mok2020,Qiu2021LearningDA,zhao2019recursive,chen2022transmorph}.
Furthermore, to allow for more flexibility in the feature extraction process or deformation modeling recent works incorporate transformer layers into the network~\citep{chen2022transmorph,chen2023transmatch,liu2022coordinate,meng2022non,wang2023modet,zhu2022swin,ghahremani2024h}.
Nonetheless, this comes at the cost of introducing high costs in terms of learnable parameters that may prove unrealistic for most real-world clinical applications outside of the registration benchmark datasets.

We argue that early-fusion approaches make inefficient use of learnable parameters, as it causes the feature extraction to learn distinct representations within the network for each possible misalignment of target-source images, increasing the task's complexity and generalizing poorly to unseen misalignments.
The black-box nature of these architectures means the extraction of intensity-derived features is inseparable from the deformation modeling process.
This poorly defined separation between the two sub-tasks of \textit{intra-domain} feature extraction and \textit{inter-domain} deformation modeling, leads to a general lack of interpretability.

An alternative approach used in literature opts for \textit{dual-stream} architectures, whereby feature extraction is performed separately from the deformation modeling process resulting in simpler and hence more interpretable inter-domain deformation modeling.
In these late-fusion works, source and target images are encoded individually prior to the concatenation of both spatial domains.
Moreover, this style of encoding produces feature hierarchies invariant to the alignment of the two spatial domains such that a change in one spatial domain has no effect on the feature extraction of the other.
Additionally, mono-modality registration can benefit in terms of parameter efficiency by reusing the same encoder across both images.

\subsection{Adaptive receptive fields and transformation refinement}

\noindent\textbf{Receptive fields and function complexity:}
In the context of deformation functions, a convolution addresses the essence of a deformation modeling task: producing a deformation vector given a neighborhood of source and target features.
When estimating the deformation at a point, the input neighborhood defines the receptive field of the operation in the source and target domain.
Although using the largest possible neighborhood sizes increases the amount of potentially relevant input information, this also increases the modeling capacity required to process the input region.
These convolutional operations can be stacked to for convolutional networks.
Deep convolutional networks offer equivalent behavior to larger neighborhood sizes proportional to their receptive fields.
At the extreme (such as with deep enough U-Nets or ViT architectures), the entirety of the spatial domain may be considered as an input argument to the deformation of any given grid point.

Conversely, having fewer neighboring input points decreases the deformation function complexity. 
While simple functions offer easily interpretable relationships between input structures and output deformations, smaller neighborhoods limit the receptive field of available structures.
A convolutional function with limited receptive field size may encounter situations where the available structural features are insufficient to capture accurate deformation. 
This becomes more prominent when the required deformations are larger than the available receptive field.
%

\noindent\textbf{Transformation refinement:}
One widely adopted technique across the literature involves progressively learning transformation through refinement steps.
By splitting the space across multiple evaluations, this technique effectively extends the available receptive field of an architecture beyond what would otherwise be available in a single prediction.
These architectures essentially implement a form of adaptive receptive fields, whereby the predicted transformation on a given source region determines the new neighborhood of target features available to the next step of the refinement process.

Arguably the most common type of refinement technique is implemented under a \textit{cascading} blueprint.
These typically involve as a series of single-stream architectures~\citep{hu2022recursive,sandkuhler2019recurrent,zhao2019recursive}, where the output transformation of an earlier process warps the source image of a later process until sufficient alignment with the target image is reached.
While these approaches demonstrate higher registration accuracy, the computational overhead increases substantially due to having to compute new transformations for the entire image at each step.
This overhead is especially evident when propagating gradients end-to-end.

\begin{figure}[h]
  \centering
  \includegraphics[width=1.\linewidth]{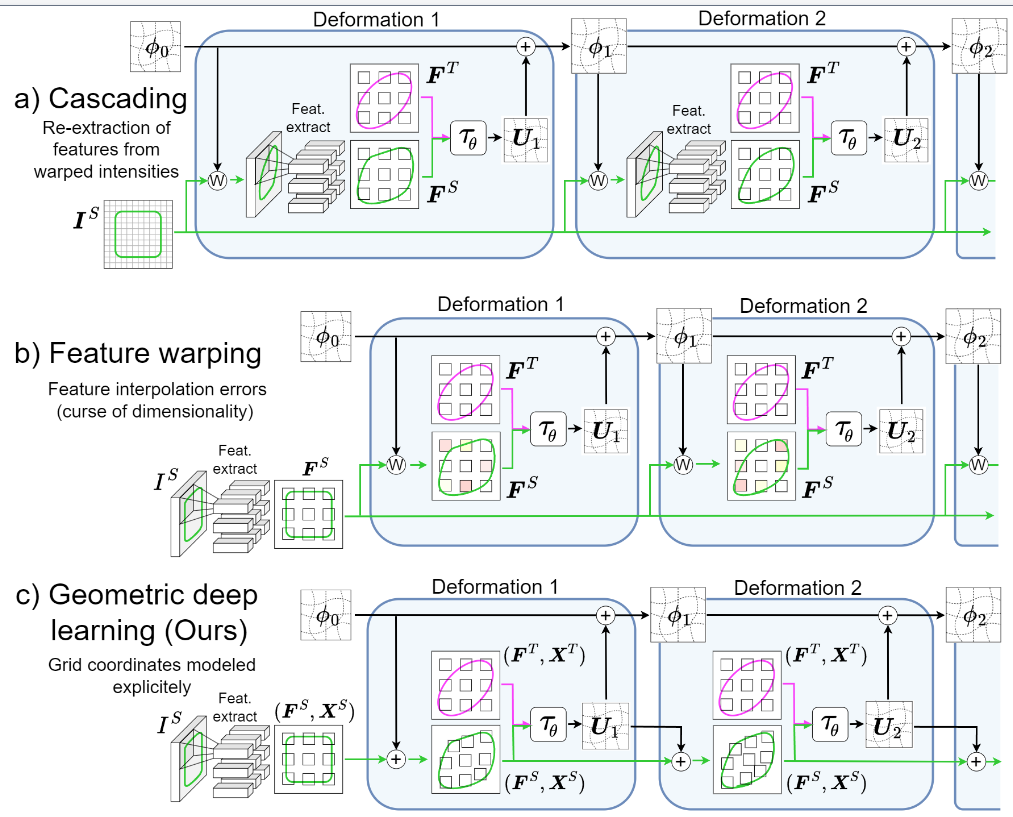}
  \caption{Different approaches to sequential deformation modeling. Warping is indicated by \textcircled{w}. a) \textbf{Cascading}: Previous transformation warps original image intensities. Modeling the next deformation requires feature re-extraction, leading to high computational costs. b) \textbf{Feature warping}: Previous transformation warps extracted source features. Computationally cheap, but warping high-dimensional features introduces interpolation errors (\textit{curse of dimensionality}). c) \textbf{Geometric deep learning}: Coordinates of features are modeled explicitly at slight memory cost. No warping is required. Deformation function \(\tau_\theta\) is aware of deformations via relative coordinates.}
  \label{fig:warping}
\end{figure} 
Another widely adopted technique of transformation refinement involves the use of \textit{multi-resolution} architectures that model transformations in a coarse-to-fine fashion.
These methods start by capturing a rough initial transformation at a coarser level and then progressively interpolating and refining at each subsequent resolution.
Multi-resolution approaches can be regarded as an extension of their single-resolution refinement counterparts, but where receptive fields are gradually shrunk throughout the refinement process (due to the increasing resolution). 
Recent works implement these coarse-to-fine architectures under a cascading blueprint~\citep{mok2020,sun2018pwc,mok2021conditional}, where the current transformation at a given resolution is used to warp the source image intensities for the next resolution.
One downside of such cascading approaches is that warped intensities need to be re-encoded prior to the next deformation prediction.

To mitigate the overhead cost associated with repeatedly encoding intensities, ~\citep{wang2023modet,ma2024iirp,kang2022dual,ghahremani2024h} propose to apply the warping operations directly on feature grids. 
Since the feature pyramids of both images are extracted before the initial transformation estimation, the input for any subsequent modeling iterations is immediately accessible through a single warping operation.
Nonetheless, the interpolation of high-dimensional vectors presents inherent challenges for the complex feature spaces in deep learning. This phenomenon, known as the \textit{curse of dimensionality}, refers to the negative impact of naive interpolation on high-dimensional representations~\citep{verleysen2005curse}.


\subsection{Limitations of grid-bounded operations and \mbox{geometric} generalization}


As spatial transformations are applied to a domain, classical grid-reliant architectures require resampling to perform further deformation predictions.
This is because their weight kernels are only defined at specified relative positions.

While the aforementioned strategy involving the warping of features alleviates the costly alternative of warping and re-encoding the original images, it nonetheless suffers from interpolation errors in high-dimensional feature spaces (\textit{curse of dimensionality}).
To mitigate these issues, this work proposes leveraging geometric deep learning (GDL), which inherently models the pattern-matching concepts of grid-based operations within a continuous domain.
This design choice obviates the aforementioned resampling issues by not being confined to a grid-based frame of reference.
Moreover, notable recent works in GDL~\cite{farahani2022deep,fuchs2020se,kashefi2021point} investigate the ability of learned functions to model the motion of sparse point-like objects.
These ideas carry striking similarities to the deformation modeling tasks involved in image registration.
Under this paradigm, points in the feature grid are defined as feature-coordinate tuples, and as such deformations imposed on these points only updates their corresponding coordinate vectors while their feature vector remains unchanged.
This offers an alternative solution to the warping trade-off in classical grid-based approaches, with the only drawback being the need to explicitly model coordinates.
Figure~\ref{fig:warping} offers a schematic highlighting the differences between these approaches.

While GDL has been successfully applied for point cloud and cortical surface registration~\citep{hansen2021deep,hansen2021graphregnet,suliman2022geomorph,shen2021accurate,hoopes2022topofit}, to the best of our knowledge, our work is the first to offer a framing of deformable image registration within the GDL paradigm.

\subsection{Contributions}

In this work, we propose a novel foundation for data-driven image registration by viewing the deformation modeling process through the lens of geometric deep learning.
While current trends call for ever-larger standardized black-box models, our formulation emphasizes how designing architectures tailored to registration tasks improves upon state-of-the-art while remaining interpretable and parameter-efficient.
We formulate our task as a coarse-to-fine process of refinement operations, where deformations are modeled via neighborhood interactions from the perspective of individual features moving in a continuous space.
This enables us to circumvent the limitations of classical grid-based deep learning operations of existing approaches.


\noindent Our contributions can be summarized as follows:
\begin{itemize}
    \item We establish a general foundation for building data-driven processes for deformable image registration tasks. Our formulation treats source and target domains as fundamentally separate coordinate systems. This allows the interaction of the two domains to be, by construction, interpretable and parameter-efficient.
    \item We frame data-driven operations under a geometric deep learning paradigm, allowing for spatially continuous input domains. When modeling sequences of transformation refinement steps, this circumvents the need for error-prone intermediate re-sampling or re-encoding operations.
    \item We demonstrate the effectiveness of our formulation by reporting improved results and robustness over current state-of-the-art deformable registration approaches.

\end{itemize}

\section{Preliminaries}
\label{sec:Preliminaries}
%
%
\noindent\textbf{Discrete data representations:}
A digitized image can be viewed as a finite grid of measurements embedded in a continuous space \(\Omega\), representing a discrete subset of intensity observations. 
We denote this representation of an image as $I = (\mathbf{I}, \mathbf{X})$ where $\mathbf{I}$ are the intensities and $\mathbf{X}$ are the discrete locations on the finite grid.
Similarly, a discrete set of features $\mathbf{F}$ extracted from an image are also embedded in the same space and can be denoted as $F = (\boldsymbol{F}, \boldsymbol{X})$. 

\noindent\textbf{Image registration:}
Given a target image \(I^T\) and a source image \(I^S\), deformable image registration (DIR) aims to find an optimal spatial transformation \(\phi^*=\arg \min_{\phi} \mathcal{J}(I^T, I^S,\phi)\), with \(\phi:\mathbbm{R}^3\rightarrow\mathbbm{R}^3\), such that the transformed source image \(I^S \circ \phi\) is most similar to the target image \(I^T\).  
Typically, this is achieved by minimizing the distance between the images with constraints on the transformation.
The overall cost \(\mathcal{J}\) is defined as:
\begin{equation}
    \mathcal{J}(I^T,I^S,\phi) = \mathcal{D}(I^T,I^S\circ\phi) + \lambda \mathcal{R}(\phi)
    \label{eq:J}
\end{equation}

\noindent where \(\mathcal{D}:\Omega \times \Omega \rightarrow \mathbbm{R}^1\) is a dissimilarity measure driving the transformation to align the images, and \(\mathcal{R}\) is a smoothness regularization on the transformation weighted by \(\lambda\). 

\noindent\textbf{Deformation function definitions:}
%
Given a pair of discrete representations of source and target images or features $(S, T)$, the deformation can be modeled using a function \(\tau: S \rightarrow \boldsymbol{U}\) that maps every point \(s \in S\) to a deformation vector \(\mathbf{u}\).
Therefore, the transformation of each discrete observation \(s\) at a coordinate \(\mathbf{x}\) is defined as \(\phi = \mathbf{x} + \mathbf{u}\).

Many data-driven methods model this deformation functions as a neural network \(\tau_\theta\) parametrized by a set of learnable weights \(\theta\).
Convolutional networks efficiently make use of their learnable weights by sharing them across the space to model a transformation at each given source point \(s \in S\).
These frameworks start by overlapping the source and target coordinate systems and predict the deformation of each point \(s\) based on only local intensities or extracted features, namely:
\begin{equation}
    \mathbf{u} = \tau\left(s, \mathcal{N}(\mathbf{x}, S), \mathcal{N}(\mathbf{x}, T) \right)
    \label{eq:tau}
\end{equation}
We use \(\mathcal{N}(\mathbf{x}, S)\) and \(\mathcal{N}(\mathbf{x}, T)\) to indicate the set of source and target points neighboring to the coordinate \(\mathbf{x}\) of source point \(s\).
In grid domains, neighborhoods are most efficiently implemented using kernel-based grid-unfolding operations.  
However, we present them in function notation to allow for any arbitrary neighborhoods around a coordinate \(\mathbf{x} \in \mathbbm{R}^3\) beyond grid domains.


In the case of a convolutional function, the concatenated input \(\left[S, T\right]\) composes the function domain, with arguments \(\mathcal{N}(\mathbf{x}, S)\) and \(\mathcal{N}(\mathbf{x}, T) \) being the \(\mathbbm{R}^{d \times k \times k \times k}\) subgrid of \(d\)-sized feature vectors available at location \(\mathbf{x}\).

\noindent\textbf{Geometric deep learning:}
Classical grid-based convolutions represent spatial patterns by modeling kernel weights at predetermined relative grid positions. 
This makes them unable to account for properties relating to the underlying coordinate system, such as variability in grid spacing or grid deformations.

In contrast, geometric deep learning (GDL) consolidates data structures (e.g., grids, cloud points) and their underlying geometric space (e.g., Euclidean, spherical) under the same mathematical framework. 
By modeling weight kernels as continuous functions \(W: \mathbbm{R}^3 \rightarrow \mathbbm{R}^d\) over a coordinate system, geometric deep learning generalizes the convolution operation for neighbors at arbitrary relative coordinates, circumventing the need for grid structures.
As such, the continuous convolution operation is described by the following equation:

\begin{equation}
   \mathbf{f}^{i} = \sum_{(\mathbf{f}^j, \mathbf{x}^j) \in \mathcal{N}(\mathbf{x}^i, F)} \mathbf{f}^j \cdot \overbrace{W\!\left(\mathbf{x}^j-\mathbf{x}^i\right)}^{\mathclap{\substack{\text{Weights evaluated at} \\ \text{a arbitrary position}}}}
    \label{eq:Graph_conv}
\end{equation}

Here, \(\mathbf{f}^{i}\) is the output feature vector for point \(i\) at location $\mathbf{x}^i$. The individual features \(\mathbf{f}^j\) in \(i\)'s neighborhood \( \mathcal{N}(\mathbf{x}^i, F)\) are aggregated by projecting them through a weighting function \(W\) based on (continuous) relative coordinates. In GDL, \(W\) is often parametrized by a neural network.

\section{Method}
\label{sec:method}
We propose a novel learning-based image registration framework named GeoReg, illustrated in Figure~\ref{fig:overview}.
Our framework follows the principle of separating feature extraction and deformation modeling as motivated in Section~\ref{sec:intro}. 
First we describe the multi-resolution feature extraction from individual images in Section~\ref{sec:method-features}.  
The extracted features are used in a novel attention-based and spatially continuous deformation refinement process, which is free from conventional grid-bounded operations and resampling errors, as detailed in Section~\ref{sec:method-refinement}. 
We combine the refinement process with a multi-resolution scheme, enhanced by a learning-based interpolation of features across different resolutions, which is presented in Section~\ref{sec:method-multires}. 
Finally in Section~\ref{sec:method-supervision}, we introduce our deep supervision formulation that provides an end-to-end signal across all refinement iterations and resolutions.

\subsection{Feature extraction}\label{sec:method-features}
We begin by creating a multi-resolution feature pyramid from source and target images respectively. 
At this stage, source and target coordinates systems are treated as separate, independent domains whose features do not interact.
We achieve this with a dual-stream encoder which employs a sequence of convolutional blocks and down-sampling operations. 
This results in target and source features at a range of fine-to-coarse resolutions \(r \in [0, 1, ..., R]\). 
Each resolution of the feature pyramid contains coarser and higher-dimensional features than its finer resolution.

During the decoding process, source and target feature grids will be overlapped into a unified coordinate system according to the current transformation. 
The deformation function \(\tau\) can then estimate deformations at each subsequent finer resolution using features of both domains. 
Note that this means the operations of \(\tau\) effectively have varying receptive field sizes over the images according to the feature resolution being used.

%
%

\subsection{Spatially continuous iterative refinement}\label{sec:method-refinement}
We formulate the deformation modeling process as an iterative refinement task. At each step of the refinement, the deformation function \(\tau\) predicts the deformation \(\mathbf{u}_n\) of a point \(s\) in the source domain at the current step \(n\) as continuous displacements.
The resulting transformation is obtained via the composition, namely \(\phi_{N}(\mathbf{x}) = \phi_{0}(\mathbf{x}) \!+\! \sum_{n=1}^{N} \!\mathbf{u}_{n}\), where $\phi_0(\mathbf{x})$ is the starting locations and $N$ is the total number of refinement steps. 
%
Each intermediate deformation $\mathbf{u}_n$ is predicted by utilizing information from neighboring points in the source domain \(\mathcal{N}(\phi_{0}(\mathbf{x}), S)\) and the target domain \(\mathcal{N}(\phi_{n}(\mathbf{x}), T)\), where \(\phi_{n}(\mathbf{x})\) is the current position of the point \(s\) being evaluated.
Note that while the neighborhood on the source domain remains fixed, the target domain neighborhood is updated dynamically during the refinement. This provides the deformation function with more information on the target domain along the path of transformation refinement, enabling us to model larger deformations beyond the limited receptive field of the neighborhood in each step.

\noindent\textbf{Spatially continuous deformation functions:}
Given that the predicted deformation $\mathbf{u}$ is continuous, the central position of the target neighborhood $\phi_n(\mathbf{x})$ is most likely floating between the original target grid points. However, the original images and features are discrete representations of data that lie on regular grids. A conventional CNN-based $\tau$ would require the neighborhood features to be re-sampled back to a regular grid.

To address this issue, we propose to use a generalized convolution under the geometric deep learning paradigm which weights neighbors based on relative positions (see Eq.~\ref{eq:Graph_conv}). 
In practice, we use the distributive property of convolution to split source and target neighborhoods into separate evaluations (see Appendix~\ref{App:DistrConvProof}). 
We implement $\tau$ using a position-aware cross-attention mechanism, namely:

\begin{align}
\mathbf{u} &= \operatorname{softmax}\left(\frac{\mathbf{q} \cdot {\mathbf{K}}^\top}{\sqrt{d}}\right) {\mathbf{V}}\\
\text{with:} \qquad \mathbf{q} &= \mathbf{f} \cdot \mathbf{W_Q} \\
\mathbf{K} &= (\mathbf{F}^{\mathcal{N}} + E\left(\mathbf{X}^{\mathcal{N}} - \phi_n(\mathbf{x}) \right)) \cdot \mathbf{W_K}\\
\mathbf{V} &= (\mathbf{F}^{\mathcal{N}} + E\left(\mathbf{X}^{\mathcal{N}} - \phi_n(\mathbf{x}) \right)) \cdot \mathbf{W_V}
\end{align}
where $\mathbf{f}$ is the feature of the source point $s$, $\mathbf{F}^{\mathcal{N}}$ and $\mathbf{X}^{\mathcal{N}}$ are the features and coordinates of the source/target neighborhood, and $\mathbf{W_Q}, \mathbf{W_K}, \mathbf{W_V}$ are learned query, key, and value matrices. 
Here \(E\) represents a positional embedding function (Fourier Features~\cite{tancik2020fourfeat}) that conditions input features with the relative coordinates of the neighboring points \(\mathbf{X}^{\mathcal{N}}\) to transformed position \(\phi_n(\mathbf{x})\) of point \(s\).
This formulation allows the deformation function $\tau$ to be repeatedly applied to refine $\phi_n(\mathbf{x})$ without re-sampling to a regular grid.
Detailed pseudocode is provided in Appendix~\ref{App:PC_tau}.


%



\subsection{Multi-resolution scheme}\label{sec:method-multires}
A further improvement in the efficiency of dynamic receptive fields can be achieved by chaining transformations across resolutions in a coarse-to-fine fashion.
By modeling deformations with features at some coarser resolutions than the original image resolution, a deformation function can capture a wider receptive field using fewer neighboring input points.
The transformation can be subsequently interpolated to the next resolution to more precisely refine deformations using narrower receptive fields.
In this way, the multi-resolution feature pyramid from the encoder allows for spatial resolution to be traded for feature expressiveness.

\noindent\textbf{Data-driven deformation interpolation:}
Taking inspiration from parametric interpolation, we define an interpolation function that produces an element-wise deformation \(\mathbf{u} = \delta^{r}\left(s, \mathcal{N}(\mathbf{x}, S^{r+1})\right)\) at a given resolution by weighting control points in the (previous) coarser resolution \(\mathcal{N}(\mathbf{x}, S^{r+1})\) nearest to a point \(s^r\) at position \(\mathbf{x}\).

We compose this cross-resolution interpolation function \(\delta\) as an initial naive inheritance step, followed by a learned interpolation component (see Figure~\ref{fig:overview}).
The initial step uses naive linear interpolation on a coarser resolution \(r\!+\!1\) of deformations \(\boldsymbol{U}^{r+1}_n\) to create a rough deformation estimate of \(\boldsymbol{U}^{r}_0\) for the next resolution of points \(S^{r}\).
This is followed by a learned refinement of the initial estimate, whereby a point \(s^{r}\) cross-attends to neighboring positionally-embedded features \(\mathcal{N}(\mathbf{x}, S^{r+1})\) in the previous resolution.
Section~\ref{App:Parametric_interp} of the Appendix further discusses the connection of learned interpolation functions to traditional parametric interpolation mechanisms. 

The reasoning behind this interpolation formulation is two-fold.
First, certain regions may require the ability to model transformations with well-defined sharp boundaries (e.g. tissues sliding along each other).
However, naive interpolation assumes smooth changes of information in the space between points.
Therefore, we introduce the learned, attention-based component to allow child points to freely refine the initial estimate based on the relevance of parent features in the coarser resolution without the smoothness prior.
Secondly, deep learning functions perform best when input and output distributions are narrowly positioned around zero.
The initial naive interpolation standardizes the distribution of child-parent relative distances around zero.

\noindent\textbf{Full decoder formulation:}
Our full decoding process alternates between deformation refinement using \(\tau\) in-resolution and interpolation using \(\delta\) cross-resolution.
This continues until the original resolution of the domain \(S\) is reached and the final deformation grid \(\boldsymbol{U}^{0}_n\) is obtained.
Detailed pseudocode is provided in Appendix~\ref{App:PC_delta}.

Generally, deep learning architectures decode information by propagating feature vectors from coarse resolutions onto finer ones.
While our architecture allows for this, we deem the coarse-to-fine propagation of features an unnecessary source of complexity.
We argue that, after aligning structures at a coarser scale, the local features present at the next resolution should be sufficient to refine deformations at that scale.
As such, our architecture does \textbf{not} carry feature vectors across decoder levels, only transformation vectors. 
We refer the reader to Appendix~\ref{App:architectural_overview} for a schematic of the overall model.

\subsection{Supervision}\label{sec:method-supervision}

\noindent\textbf{Iterative refinement supervision:}
We define the iterative refinement objective as follows:
\begin{equation}
    \mathcal{J}_{\text{refine}}
    = \sum_{n=1}^{N}\mathcal{J}(I^T,I^S, \phi_{n})
    \label{Eq:Res_objective}
\end{equation}
\noindent where \(N\) is the maximum iteration steps and \(\phi_{n}\) is the transformation after the \(n\)-th refinement prediction. 
By supervising each transformation of the refinement process, this objective encourages intermediate steps to find early meaningful correspondances.

\noindent\textbf{Cross-resolution supervision:}
A downside of predicting the deformation grid \(\boldsymbol{U}^r\) at some coarser resolution \(r\! >\! 0\) is that dissimilarity metrics require the original intensities be downsampled to match the resolution of the deformation grid. 
The absence of high-frequency information during supervision risks finding correspondences based on averaged intensity values that could potentially pose a poor match at the original image resolution.
We thus propose to extend the registration objective in Eq.~\ref{Eq:Res_objective} by modeling deformations at coarser resolution where receptive fields are large, while supervising at a finer resolution where spatial resolution most benefits dissimilarity metrics.

We compute the objective at the finer resolution \(r\) using the transformation \(\phi^{r}_{0} = \boldsymbol{X}^r + \boldsymbol{U}^{r}_0\).
This will propagate supervision gradients directly through \(\delta^{r}\) to the deformations predicted by the deformation function \(\tau^{r+1}\) of the previous resolution:

\begin{equation}
    \mathcal{J_{\text{interp}}} 
    = \mathcal{J}(I^T, I^S, \phi^{r}_{0})
    \label{Eq:Cross_Res_objective}
\end{equation}

\noindent\textbf{Full multi-resolution objective:}
The objectives naturally extend to any number of resolutions by supervising the intermediate output transformation of every interpolation and refinement step up to the original image resolution. 
This results in a formulation where the deformation predicted for a source point \(s^r\) at some coarser resolution \(r\! >\! 0\) is responsible for the majority of the transformation modeled at that region of space.
This is because the deformations \([\mathbf{u}_0, ..., \mathbf{u}_N]\) applied to \(s^r\) are supervised via all the down-stream child points of finer resolutions \([r\!-\!1, ..., 0]\) that use the interpolated deformation of \(s^r\) as a basis to further optimize their dissimilarities.
%
%
Formally, the full multi-resolution objective can be written as:
\begin{equation}
    \mathcal{J_{\text{multi-res}}}
    = \mathcal{J}^{R}_{\text{refine}} + \sum_{r=0}^{R-1} \alpha^{r}\left(\mathcal{J}^{r}_{\text{interp}}  + \mathcal{J}^{r}_{\text{refine}} \right)
    \label{Eq:Multi_Res_objective}
\end{equation}

\noindent where \(\mathcal{J}^{r}_{\text{interp}}\) and \(\mathcal{J}^{r}_{\text{refine}}\) are the \(r\)-th resolution interpolation and refinement losses defined in Eq.~\ref{Eq:Res_objective} and Eq.~\ref{Eq:Cross_Res_objective}, where appropriate downsampling is applied to source and target intensities to compute dissimilarity.
We incorporate a resolution-specific weighting factor \(\alpha\) as a general term to weigh dissimilarity and regularization components at depending on resolution. 

\section{Experiments}

\subsection{Illustration of GeoReg properties}
In this section, we investigate the qualitative properties of our approach using a small version of our architecture.
We perform a same-digit registration task on the MNIST dataset~\cite{deng2012mnist}.
We encode at three resolutions [28x28, 14x14, 7x7], with feature sizes [16, 32, 64], followed by 3x3 decoder neighborhoods at each resolution.

\begin{figure}[h]
  \centering
  \includegraphics[width=\linewidth]{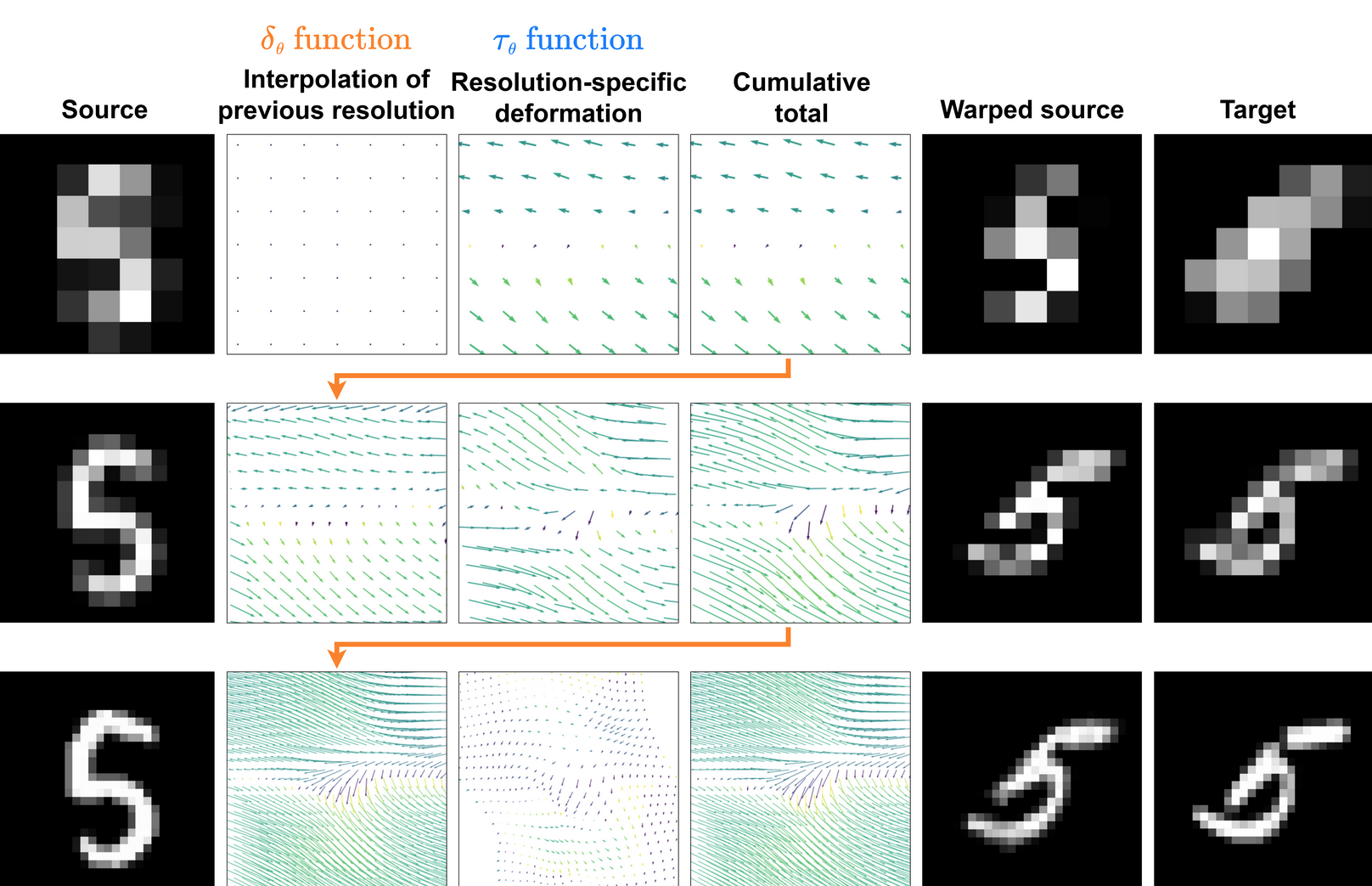}
  \caption{Visualization of the registration process of an MNIST image pair over 3 resolutions. The multi-resolution approach naturally gives rise to deformation structures and magnitudes depending on the scale.}
  \label{fig:Qualitative_panels}
\end{figure}

\noindent\textbf{Scale separation:} Figure~\ref{fig:Qualitative_panels} displays the intermediate deformation vectors of a registration process between a pair of digits.
We observe a well-defined separation in what types of deformations our architecture models across resolutions.
At the coarsest resolution, the largest components of the transformation are captured, such as rotation or shearing of various parts of the image.
Meanwhile, the deformations modeled at the middle resolution appear to correspond to local morphological differences in the shapes of the digits.
At the finest resolution, with the transformation predominantly captured by previous resolutions, only sub-pixel adjustments are modeled. 
This scale separation allows different regularizations and similarity metrics at each resolution depending on the intended application of a user.

Furthermore, the multi-resolution objective offers a powerful form of implicit regularization to a transformation. 
Although no visual features are present in the backgrounds of these digits, the multi-resolution formulation generalizes a transformation to these regions based on deformations at coarser levels.
We believe this to be an important behavior for medical imaging, where large featureless regions may need to be guided by coarser resolutions that have access to larger receptive fields.

\begin{figure}[h]
  \centering
  \includegraphics[width=\linewidth]{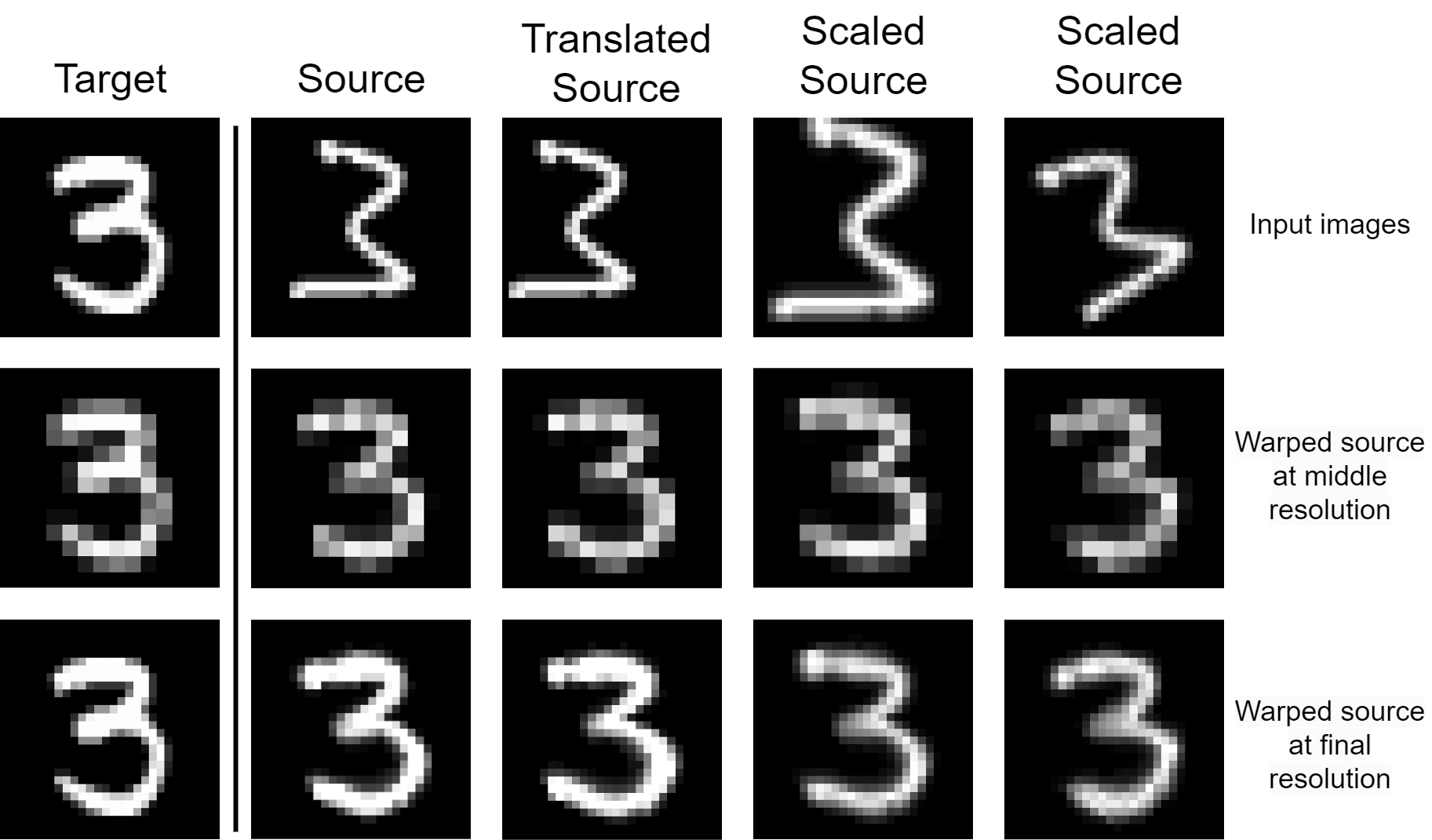}
  \caption{Visualization of the registration process of an MNIST image pair under various source image augmentations. Early through the multi-resolution process, the model appears to remove most variation across augmented instances.}
  \label{fig:Attention_panels}
\end{figure}

\noindent\textbf{Alignment at coarser levels:} 
To validate the hypothesis that our approach predominantly models the largest components of the transformation at the coarser resolutions, we investigate intermediate transformations under multiple augmentations of the same source image.
Figure~\ref{fig:Attention_panels} shows how at coarser scales, most of the variation introduced by augmentations on the source domain is no longer present. 
From the perspective of the last resolution layer, all augmented source instances down to the same structural alignment.
This substantially simplifies the remaining modeling task undertaken by later decoder levels.

\subsection{Results on medical datasets}
\label{sec:results}
\begin{figure*}[h]
  \centering
  \includegraphics[width=1\textwidth]{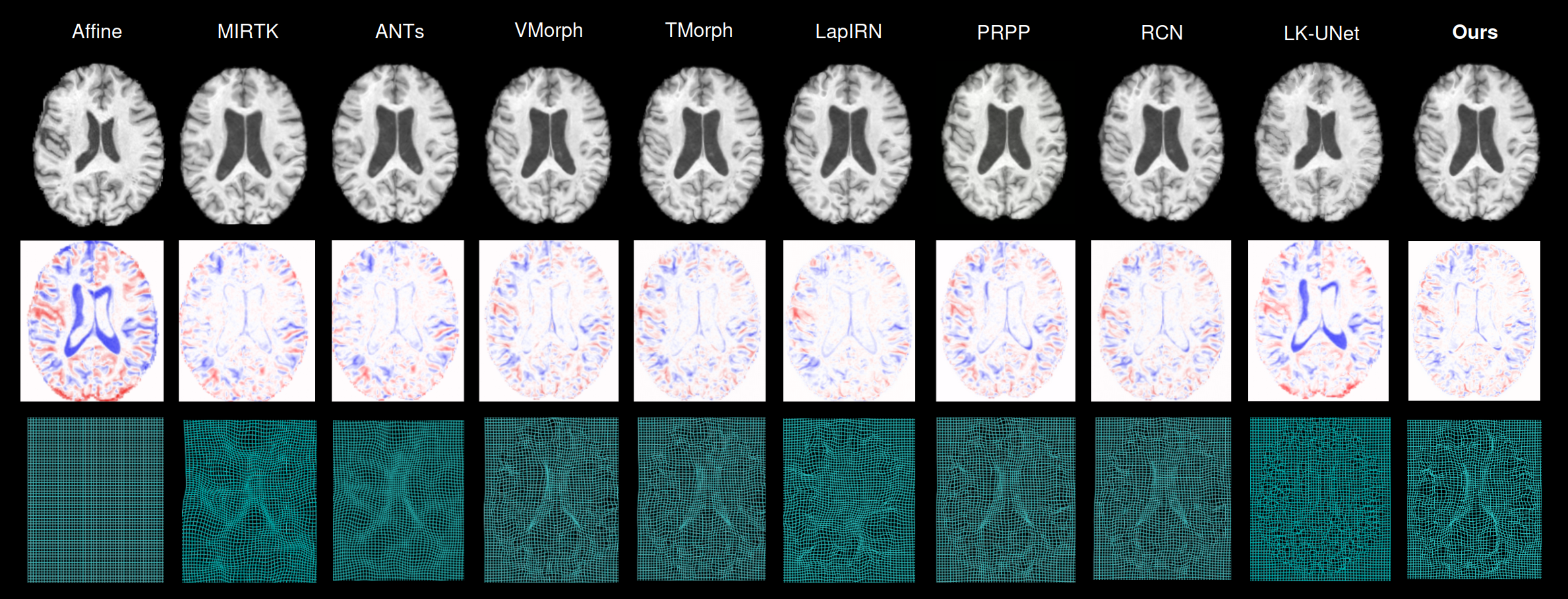}
  \caption{Qualitative results of all compared methods for the CamCAN T1w-T1w inter-subject deformable registration experiment.}
  \label{fig:qualitative_Results}
  \end{figure*}

\begin{table*}[h]
    \small
    \addtolength{\tabcolsep}{4pt}
    \centering
    \caption{Quantitative results measuring the accuracy and regularity of different registration methods on brain T1w inter-subject (CamCAN) and intra-subject longitudinal retinal optical coherence tomography (OCT) dataset registration. The performance of GeoReg with bilinear feature warping instead of a learned interpolation component $\delta$ is shown under `feat. warp'.
    }
    \label{tab:hyperparameters}
    \label{tab:quant_def}
    \setlength{\tabcolsep}{3pt}
    \resizebox{\textwidth}{!}{
    \begin{tabular}{l|c|ccc|ccc|ccc}
    \toprule
    & & \multicolumn{3}{c|}{\textbf{Brain CamCAN}} & \multicolumn{3}{c|}{\textbf{Brain CamCAN T1T2}} & \multicolumn{3}{c}{\textbf{OCT}}\\
    \toprule
    & \# Param. & DSC \(\uparrow\) & HD95 \(\downarrow\) & Folding (\%)  \(\downarrow\) & DSC \(\uparrow\) & HD95 \(\downarrow\) & Folding (\%)  \(\downarrow\) & DSC \(\uparrow\) & HD95 \(\downarrow\) & Folding (\%)  \(\downarrow\) \\
    \midrule
    Affine                      & -          & \(0.618 \pm 0.01\)          & \(5.46\pm 0.07\)           & -                                & \(0.618 \pm 0.49\)          & \(5.46\pm 1.44\)  & -                               & \(0.368 \pm 0.14\)                   & \(4.74\pm 3.50\)                    & -                                                                         \\
    ANTs [SyN]              & -          & \(0.808 \pm 0.06\)          & \(2.55 \pm 0.08\)                    & \(0.15 \pm 0.052\)  & \(0.767 \pm 0.02\)                    & \(3.05 \pm 0.90\)           & \(0.10 \pm 0.01\) & \(0.512 \pm 0.02\) & \(2.09 \pm 0.12\) & \(0.008\pm 0.002\) \\
    MIRTK [FFD]             & -          & \(0.813 \pm 0.07\)          & \(1.97 \pm 0.54\)                    & \(0.21 \pm 0.057\)  & \(0.768 \pm 0.01\)                    & \(2.23 \pm 0.44\)           & \(0.10 \pm 0.01\) & \(0.512 \pm 0.02\) & \(2.09 \pm 0.12\) & \(0.008 \pm 0.002\) \\
   VoxelMorph                  & \(320\) k  & \(0.800 \pm 0.02\)          & \(2.38 \pm 0.74\)          & \(0.02 \pm 0.006\) & \(0.756 \pm 0.03\)          & \(3.83 \pm 1.27\) & \(0.23 \pm 0.09\) & \(0.566 \pm 0.08\)                             & \(2.03 \pm2.47\)                    & \(0.001 \pm 0.001\)                                         \\
   LapIRN                      & \(924\) k  & \(0.820 \pm 0.04\)          & \(3.06 \pm 1.40\)          & \(0.31 \pm 0.002\)  & \(0.784 \pm 0.06\)          & \(\mathbf{2.18 \pm 0.45}\) & \(0.23 \pm 0.03\) & {\(0.516 \pm 0.28\)}             & {\(2.86 \pm 0.18\)}             & {\(0.005 \pm 0.007\)}                                     \\
Transmorph                  & \(46.8\) M & \(0.822 \pm 0.01\)          & \(1.92 \pm 0.52\)          & \(0.02\pm 0.005\)  & \(\mathbf{0.788 \pm 0.02}\)                    & \(2.96\pm 1.00\)            & \(0.32 \pm 0.01\) & \(0.577 \pm 0.12\)                             & \(1.95 \pm 2.54\)                   & \(0.004 \pm 0.033\)                                         \\
D-PRNet                     & \(1.2\) M  & \(0.795\pm 0.01\)           & \(2.52 \pm 0.91\)          & \(0.02 \pm 0.007\) & \(0.745 \pm 0.03\)          & \(4.42 \pm 1.21\) & \(0.28 \pm 0.07\) & \(0.562 \pm 0.11\)                   & \(2.67 \pm 3.19\)                   & \(0.001 \pm 0.001\)                                         \\
RCN                         & \(282\) M  & \(0.818 \pm 0.01\)          & \(\mathbf{1.92 \pm 0.55}\)                    & \(0.01 \pm 0.004\) & \(0.776 \pm 0.14\)                    & \(2.58 \pm 0.54\)           & \(0.25 \pm 0.06\) & \(0.562\pm 0.07\)                              & \(1.89\pm 2.33\)                    & \(0.009 \pm 0.003\)                                         \\
LK-UNet                     & \(1.1\) M  & \(0.727 \pm 0.04\)          & \(4.32 \pm 1.53\)                    & \(0.01 \pm 0.002\) & \(0.697 \pm 0.04\)                    & \(4.94 \pm 1.16\)           & \(0.25 \pm 0.08\) & \(0.438 \pm 0.15\)                   & \(4.18 \pm 3.52\)                    & \(0.007\pm 0.001\)                                          \\
\textbf{Ours} (feat. warp) & \(657\) k  & \(0.801 \pm 0.07\)          & \(2.51 \pm 0.95\)                    & \(0.09 \pm 0.009\)  & \(0.727 \pm 0.07\)          & \(2.95 \pm 1.16\)  & \(0.27 \pm 0.03\) & \(0.560 \pm 0.07\)                   & \(2.01 \pm 2.27\)          & \(0.008 \pm 0.018\)                                           \\
\midrule
\textbf{Ours} (GeoReg)     & \(741\) k  & \(\mathbf{0.838 \pm 0.06}\) & \(2.05 \pm 0.82\) & \(0.09 \pm 0.006\)  & \(\mathbf{0.788 \pm 0.05}\) & \(2.53 \pm 1.45\) & \(0.26 \pm 0.08\) & \(\mathbf{0.581\pm 0.09}\)                    & \(\mathbf{1.72 \pm 1.72}\)          & \(0.003 \pm 0.019\)                                           \\
\bottomrule 
    \end{tabular}
}

\end{table*}


To assess the capability of our method to recover deformable transformations, we conduct a comparative evaluation against several widely used baseline methods in three distinct tasks. 
We evaluate our method on inter-subject registration of T1w-T1w MRI brain images, as well as the more challenging multi-contrast T1w-T2w brain images from the CamCAN dataset~\citep{Shafto2014TheCC,Taylor2017TheCC}. 

Moreover, we perform longitudinal intra-subject registration using the retinal optical coherence tomography (OCT)~\cite{sutton2023developing} dataset. 
This dataset presents significant challenges including substantial noise in the images, large misalignments in the position of the retina and substantial retina deterioration for acquisitions further apart in time.
Further details on datasets and pre-processing may be found in Appendix~\ref{App:data_info}.
We refer the reader to Appendix sections~\ref{App:Baselines} and~\ref{App:implementation_details} for baselines and implementation details.

The results presented in Table~\ref{tab:hyperparameters} demonstrate our method outperforms others on the challenging longitudinal OCT registration task, exhibiting robust performance in the presence of noise and significant misalignments across time.
On both mono- and multi-modal inter-subject brain registration tasks, our approach shows on par performance with the state-of-the-art methods.
For a qualitative inspection of the registration results, we refer to Fig. \ref{fig:qualitative_Results} and Fig.~\ref{App:fig:b_qual1}-\ref{App:fig:oct_qual} of the Appendix.

\section{Conclusion}
\label{sec:conclusion}

In this work, we introduce a novel formulation of deformable image registration by using geometric deep-learning principles. 
We discuss the benefits of estimating deformations on non-fixed grid locations by defining data-driven functions on continuous domains.
We outline the need for two types of learned continuous operations: A deformation modeling function \(\tau\) and a cross-resolution interpolation function \(\delta\). 
Our model outperforms various deformable registration baselines on challenging OCT deformable registration tasks.
On an MRI brain registration task, our approach shows performance on par with state-of-the-art methods.

We think that this contribution opens up avenues of research to reduce the black-box nature of current learned registration paradigms and incorporate ideas from conventional image registration into deep learning architectures.
%
%
%
Despite optimizations on many aspects of the data represented in memory, explicitly modeling coordinates causes our method to have a larger memory footprint than its grid-reliant counterparts.
Nonetheless, the ability of our deformation functions to directly incorporate volume spacing into the deformation prediction presents an interesting avenue to overcome the limitations of current registration approaches in anisotropic spacing tasks.

{
    \small
    \bibliographystyle{ieeenat_fullname}
    \bibliography{main}

\begin{thebibliography}{45}
\providecommand{\natexlab}[1]{#1}
\providecommand{\url}[1]{\texttt{#1}}
\expandafter\ifx\csname urlstyle\endcsname\relax
  \providecommand{\doi}[1]{doi: #1}\else
  \providecommand{\doi}{doi: \begingroup \urlstyle{rm}\Url}\fi

\bibitem[Avants et~al.(2008)Avants, Epstein, Grossman, and Gee]{avants2008symmetric}
B.~B. Avants, C.~L. Epstein, M. Grossman, and J.~C. Gee.
\newblock Symmetric diffeomorphic image registration with cross-correlation: evaluating automated labeling of elderly and neurodegenerative brain.
\newblock \emph{Medical image analysis}, 12\penalty0 (1):\penalty0 26--41, 2008.

\bibitem[Avants et~al.(2009)Avants, Tustison, Song, et~al.]{avants2009advanced}
B.~B. Avants, N. Tustison, G. Song, et~al.
\newblock Advanced normalization tools ({ANTS}).
\newblock \emph{Insight j}, 2\penalty0 (365):\penalty0 1--35, 2009.

\bibitem[Balakrishnan et~al.(2019)Balakrishnan, Zhao, Sabuncu, Guttag, and Dalca]{balakrishnan2019}
G. Balakrishnan, A. Zhao, M.~R. Sabuncu, J. Guttag, and A.~V. Dalca.
\newblock Voxelmorph: a learning framework for deformable medical image registration.
\newblock \emph{IEEE Transactions on Medical Imaging}, 38\penalty0 (8):\penalty0 1788--1800, 2019.

\bibitem[Bronstein et~al.(2021)Bronstein, Bruna, Cohen, and Veli{\v{c}}kovi{\'c}]{bronstein2021geometric}
M.~M. Bronstein, J. Bruna, T. Cohen, and P. Veli{\v{c}}kovi{\'c}.
\newblock Geometric deep learning: Grids, groups, graphs, geodesics, and gauges.
\newblock \emph{arXiv preprint arXiv:2104.13478}, 2021.

\bibitem[Chen et~al.(2022)Chen, Frey, He, Segars, Li, and Du]{chen2022transmorph}
J. Chen, E.~C. Frey, Y. He, W.~P. Segars, Y. Li, and Y. Du.
\newblock Transmorph: Transformer for unsupervised medical image registration.
\newblock \emph{Medical image analysis}, 82:\penalty0 102615, 2022.

\bibitem[Chen et~al.(2023)Chen, Zheng, and Gee]{chen2023transmatch}
Z. Chen, Y. Zheng, and J.~C. Gee.
\newblock Transmatch: a transformer-based multilevel dual-stream feature matching network for unsupervised deformable image registration.
\newblock \emph{IEEE Transactions on Medical Imaging}, 2023.

\bibitem[Dalca et~al.(2018)Dalca, Balakrishnan, Guttag, and Sabuncu]{Dalca2018UnsupervisedLF}
A.~V. Dalca, G. Balakrishnan, J.~V. Guttag, and M.~R. Sabuncu.
\newblock Unsupervised learning for fast probabilistic diffeomorphic registration.
\newblock In \emph{International Conference on Medical Image Computing and Computer-Assisted Intervention}, 2018.

\bibitem[Deng(2012)]{deng2012mnist}
L. Deng.
\newblock The mnist database of handwritten digit images for machine learning research [best of the web].
\newblock \emph{IEEE signal processing magazine}, 29\penalty0 (6):\penalty0 141--142, 2012.

\bibitem[Farahani and Hamker(2022)]{farahani2022deep}
J. Farahani, A.and~Vitay and F.~H. Hamker.
\newblock Deep neural networks for geometric shape deformation.
\newblock In \emph{German Conference on Artificial Intelligence (K{\"u}nstliche Intelligenz)}, pages 90--95. Springer, 2022.

\bibitem[Fuchs et~al.(2020)Fuchs, Worrall, Fischer, and Welling]{fuchs2020se}
F. Fuchs, D. Worrall, V. Fischer, and M. Welling.
\newblock Se (3)-transformers: 3d roto-translation equivariant attention networks.
\newblock \emph{Advances in neural information processing systems}, 33:\penalty0 1970--1981, 2020.

\bibitem[Ghahremani et~al.(2024)Ghahremani, Khateri, Jian, Wiestler, Adeli, and Wachinger]{ghahremani2024h}
Morteza Ghahremani, Mohammad Khateri, Bailiang Jian, Benedikt Wiestler, Ehsan Adeli, and Christian Wachinger.
\newblock H-{V}i{T}: A hierarchical vision transformer for deformable image registration.
\newblock In \emph{Proceedings of the IEEE/CVF Conference on Computer Vision and Pattern Recognition}, pages 11513--11523, 2024.

\bibitem[Hansen and Heinrich(2021{\natexlab{a}})]{hansen2021deep}
L. Hansen and M.~P. Heinrich.
\newblock Deep learning based geometric registration for medical images: How accurate can we get without visual features?
\newblock In \emph{Information Processing in Medical Imaging: 27th International Conference, IPMI 2021, Virtual Event, June 28--June 30, 2021, Proceedings 27}, pages 18--30. Springer, 2021{\natexlab{a}}.

\bibitem[Hansen and Heinrich(2021{\natexlab{b}})]{hansen2021graphregnet}
L. Hansen and M.~P. Heinrich.
\newblock Graphregnet: Deep graph regularisation networks on sparse keypoints for dense registration of {3D} lung {CTs}.
\newblock \emph{IEEE Transactions on Medical Imaging}, 40\penalty0 (9):\penalty0 2246--2257, 2021{\natexlab{b}}.

\bibitem[Haskins et~al.(2020)Haskins, Kruger, and Yan]{haskins2020}
G. Haskins, U. Kruger, and P. Yan.
\newblock Deep learning in medical image registration: a survey.
\newblock \emph{Machine Vision and Applications}, 31:\penalty0 1--18, 2020.

\bibitem[Hoopes et~al.(2022)Hoopes, Iglesias, Fischl, Greve, and Dalca]{hoopes2022topofit}
A. Hoopes, J.~E. Iglesias, B. Fischl, D. Greve, and A.~V. Dalca.
\newblock {TopoFit}: rapid reconstruction of topologically-correct cortical surfaces.
\newblock \emph{Proceedings of machine learning research}, 172:\penalty0 508, 2022.

\bibitem[Horn(2016)]{Horn2016}
A. Horn.
\newblock {MNI T1 6thGen NLIN to MNI 2009b NLIN ANTs transform}.
\newblock 2016.

\bibitem[Hu et~al.(2022)Hu, Zhou, Xiong, and Wu]{hu2022recursive}
B. Hu, S. Zhou, Z. Xiong, and F. Wu.
\newblock Recursive decomposition network for deformable image registration.
\newblock \emph{IEEE Journal of Biomedical and Health Informatics}, 26\penalty0 (10):\penalty0 5130--5141, 2022.

\bibitem[Iglesias et~al.(2011)Iglesias, Liu, Thompson, and Tu]{Iglesias2011Robex}
J. Iglesias, C. Liu, P. Thompson, and Z. Tu.
\newblock Robust brain extraction across datasets and comparison with publicly available methods.
\newblock \emph{IEEE Transactions on Medical Imaging}, 30(9):\penalty0 1617--1634, 2011.

\bibitem[Jia et~al.(2022)Jia, Bartlett, Zhang, Lu, Qiu, and Duan]{jia2022u}
X. Jia, J. Bartlett, T. Zhang, W. Lu, Z. Qiu, and J. Duan.
\newblock U-net vs transformer: Is {u}-net outdated in medical image registration?
\newblock In \emph{International Workshop on Machine Learning in Medical Imaging}, pages 151--160. Springer, 2022.

\bibitem[Kang et~al.(2022)Kang, Hu, Huang, Scott, and Reyes]{kang2022dual}
M. Kang, X. Hu, W. Huang, M.~R. Scott, and M. Reyes.
\newblock Dual-stream pyramid registration network.
\newblock \emph{Medical image analysis}, 78:\penalty0 102379, 2022.

\bibitem[Kashefi et~al.(2021)Kashefi, Rempe, and Guibas]{kashefi2021point}
A. Kashefi, D. Rempe, and L.s~J. Guibas.
\newblock A point-cloud deep learning framework for prediction of fluid flow fields on irregular geometries.
\newblock \emph{Physics of Fluids}, 33\penalty0 (2), 2021.

\bibitem[Ledig et~al.(2015)Ledig, Heckemann, Hammers, L{\'o}pez, Newcombe, Makropoulos, L{\"o}tj{\"o}nen, Menon, and Rueckert]{Ledig2015RobustWS}
C. Ledig, R. Heckemann, A. Hammers, J. L{\'o}pez, V. Newcombe, A. Makropoulos, J. L{\"o}tj{\"o}nen, D. Menon, and D. Rueckert.
\newblock Robust whole-brain segmentation: Application to traumatic brain injury.
\newblock \emph{Medical image analysis}, 21 1:\penalty0 40--58, 2015.

\bibitem[Liu et~al.(2022)Liu, Zuo, Han, Xue, Prince, and Carass]{liu2022coordinate}
Y. Liu, L. Zuo, S. Han, Y. Xue, J.~L. Prince, and A. Carass.
\newblock Coordinate translator for learning deformable medical image registration.
\newblock In \emph{International Workshop on Multiscale Multimodal Medical Imaging}, pages 98--109. Springer, 2022.

\bibitem[Lowekamp et~al.(2013)Lowekamp, Chen, Ib{\'a}{\~n}ez, and Blezek]{Lowekamp2013TheDO}
B. Lowekamp, D. Chen, L. Ib{\'a}{\~n}ez, and D. Blezek.
\newblock The design of simpleitk.
\newblock \emph{Frontiers in Neuroinformatics}, 7, 2013.

\bibitem[Ma et~al.(2024)Ma, Zhang, Li, and Wen]{ma2024iirp}
Tai Ma, Suwei Zhang, Jiafeng Li, and Ying Wen.
\newblock {IIRP-Net}: iterative inference residual pyramid network for enhanced image registration.
\newblock In \emph{Proceedings of the IEEE/CVF Conference on Computer Vision and Pattern Recognition}, pages 11546--11555, 2024.

\bibitem[Meng et~al.(2022)Meng, Bi, Feng, and Kim]{meng2022non}
M. Meng, L. Bi, D. Feng, and J. Kim.
\newblock Non-iterative coarse-to-fine registration based on single-pass deep cumulative learning.
\newblock In \emph{International Conference on Medical Image Computing and Computer-Assisted Intervention}, pages 88--97. Springer, 2022.

\bibitem[Mok and Chung(2021)]{mok2021conditional}
T.C.W. Mok and A.C.S. Chung.
\newblock Conditional deformable image registration with convolutional neural network.
\newblock In \emph{Medical Image Computing and Computer Assisted Intervention--MICCAI 2021: 24th International Conference, Strasbourg, France, September 27--October 1, 2021, Proceedings, Part IV 24}, pages 35--45. Springer, 2021.

\bibitem[Mok and Chung(2020)]{mok2020}
T.~C.~W. Mok and A.~C.~S. Chung.
\newblock Large deformation diffeomorphic image registration with {Laplacian} pyramid networks.
\newblock In \emph{Medical Image Computing and Computer Assisted Intervention--MICCAI 2020: 23rd International Conference, Lima, Peru, October 4--8, 2020, Proceedings, Part III 23}, pages 211--221. Springer, 2020.

\bibitem[Qiu et~al.(2021)Qiu, Qin, Schuh, Hammernik, and Rueckert]{Qiu2021LearningDA}
H. Qiu, C. Qin, A. Schuh, K. Hammernik, and D. Rueckert.
\newblock Learning diffeomorphic and modality-invariant registration using {B}-splines.
\newblock In \emph{International Conference on Medical Imaging with Deep Learning}, 2021.

\bibitem[Rueckert et~al.(1999)Rueckert, Sonoda, Hayes, Hill, Leach, and Hawkes]{Rueckert1999NonrigidRU}
D. Rueckert, L.~I. Sonoda, C. Hayes, D.~L.~G. Hill, M.~O. Leach, and D.~J. Hawkes.
\newblock Nonrigid registration using free-form deformations: application to breast {MR} images.
\newblock \emph{IEEE Transactions on Medical Imaging}, 18:\penalty0 712--721, 1999.

\bibitem[Sandk{\"u}hler et~al.(2019)Sandk{\"u}hler, Andermatt, Bauman, Nyilas, Jud, and Cattin]{sandkuhler2019recurrent}
R. Sandk{\"u}hler, S. Andermatt, G. Bauman, S. Nyilas, C. Jud, and P.~C. Cattin.
\newblock Recurrent registration neural networks for deformable image registration.
\newblock \emph{Advances in Neural Information Processing Systems}, 32, 2019.

\bibitem[Schuh et~al.(2014)Schuh, Murgasova, Makropoulos, Ledig, Counsell, Hajnal, Aljabar, and Rueckert]{Schuh2014ConstructionOA}
A. Schuh, M. Murgasova, A. Makropoulos, C. Ledig, S. Counsell, J. Hajnal, P. Aljabar, and D. Rueckert.
\newblock Construction of a 4{D} brain atlas and growth model using diffeomorphic registration.
\newblock In \emph{STIA}, 2014.

\bibitem[Shafto et~al.(2014)Shafto, Tyler, Dixon, Taylor, Rowe, Cusack, Calder, Marslen-Wilson, Duncan, Dalgleish, Henson, Brayne, and Matthews]{Shafto2014TheCC}
M. Shafto, L. Tyler, M. Dixon, Jason~R. Taylor, J. Rowe, R. Cusack, A. Calder, W.~D. Marslen-Wilson, J. Duncan, T. Dalgleish, R. Henson, C. Brayne, and F. Matthews.
\newblock The {Cambridge} centre for ageing and neuroscience {(Cam-CAN)} study protocol: a cross-sectional, lifespan, multidisciplinary examination of healthy cognitive ageing.
\newblock \emph{BMC Neurology}, 14, 2014.

\bibitem[Shen et~al.(2021)Shen, Feydy, Liu, Curiale, San Jose~Estepar, San Jose~Estepar, and Niethammer]{shen2021accurate}
Z. Shen, J. Feydy, P. Liu, A.~H. Curiale, R. San Jose~Estepar, R. San Jose~Estepar, and M. Niethammer.
\newblock Accurate point cloud registration with robust optimal transport.
\newblock \emph{Advances in Neural Information Processing Systems}, 34:\penalty0 5373--5389, 2021.

\bibitem[Sotiras et~al.(2013)Sotiras, Davatzikos, and Paragios]{Sotiras2013DeformableMI}
A. Sotiras, C. Davatzikos, and N. Paragios.
\newblock Deformable medical image registration: A survey.
\newblock \emph{IEEE Transactions on Medical Imaging}, 32:\penalty0 1153--1190, 2013.

\bibitem[Suliman et~al.(2022)Suliman, Williams, Fawaz, and Robinson]{suliman2022geomorph}
M.~A. Suliman, L.~Z.~J. Williams, A. Fawaz, and E.~C. Robinson.
\newblock Geomorph: Geometric deep learning for cortical surface registration.
\newblock In \emph{Geometric Deep Learning in Medical Image Analysis}, 2022.

\bibitem[Sun et~al.(2018)Sun, Yang, Liu, and Kautz]{sun2018pwc}
D. Sun, X. Yang, M.-Y. Liu, and J. Kautz.
\newblock {Pwc-net: Cnns} for optical flow using pyramid, warping, and cost volume.
\newblock In \emph{Proceedings of the IEEE conference on computer vision and pattern recognition}, pages 8934--8943, 2018.

\bibitem[Sutton et~al.(2023)Sutton, Menten, Riedl, Bogunovi{\'c}, Leingang, Anders, Hagag, Waldstein, Wilson, Cree, et~al.]{sutton2023developing}
J. Sutton, M.~J. Menten, S. Riedl, H. Bogunovi{\'c}, O. Leingang, P. Anders, A.~M. Hagag, S. Waldstein, A. Wilson, A.~J. Cree, et~al.
\newblock Developing and validating a multivariable prediction model which predicts progression of intermediate to late age-related macular degeneration—the {PINNACLE} trial protocol.
\newblock \emph{Eye}, 37\penalty0 (6):\penalty0 1275--1283, 2023.

\bibitem[Tancik et~al.(2020)Tancik, Srinivasan, Mildenhall, Fridovich-Keil, Raghavan, Singhal, Ramamoorthi, Barron, and Ng]{tancik2020fourfeat}
M. Tancik, P. Srinivasan, B. Mildenhall, S. Fridovich-Keil, N. Raghavan, U. Singhal, R. Ramamoorthi, J. Barron, and R. Ng.
\newblock Fourier features let networks learn high frequency functions in low dimensional domains.
\newblock \emph{NeurIPS}, 2020.

\bibitem[Taylor et~al.(2017)Taylor, Williams, Cusack, Auer, Shafto, Dixon, Tyler, Group, and Henson]{Taylor2017TheCC}
J. Taylor, N. Williams, R. Cusack, T. Auer, M. Shafto, M. Dixon, L. Tyler, Cam-CAN Group, and R. Henson.
\newblock The cambridge centre for ageing and neuroscience {(Cam-CAN)} data repository: Structural and functional {MRI}, {MEG}, and cognitive data from a cross-sectional adult lifespan sample.
\newblock \emph{Neuroimage}, 144:\penalty0 262 -- 269, 2017.

\bibitem[Verleysen and Fran{\c{c}}ois(2005)]{verleysen2005curse}
M. Verleysen and D. Fran{\c{c}}ois.
\newblock The curse of dimensionality in data mining and time series prediction.
\newblock In \emph{International work-conference on artificial neural networks}, pages 758--770. Springer, 2005.

\bibitem[Wang et~al.(2023)Wang, Ni, and Wang]{wang2023modet}
H. Wang, D. Ni, and Y. Wang.
\newblock Modet: Learning deformable image registration via motion decomposition transformer.
\newblock In \emph{International Conference on Medical Image Computing and Computer-Assisted Intervention}, pages 740--749. Springer, 2023.

\bibitem[Xiao et~al.(2021)Xiao, Teng, Liu, Li, Ren, Yang, Shen, and Cai]{xiao2021review}
H. Xiao, X. Teng, C. Liu, T. Li, G. Ren, R. Yang, D. Shen, and J. Cai.
\newblock A review of deep learning-based three-dimensional medical image registration methods.
\newblock \emph{Quantitative Imaging in Medicine and Surgery}, 11\penalty0 (12):\penalty0 4895, 2021.

\bibitem[Zhao et~al.(2019)Zhao, Dong, Chang, Xu, et~al.]{zhao2019recursive}
S. Zhao, Y. Dong, E.~I. Chang, Y. Xu, et~al.
\newblock Recursive cascaded networks for unsupervised medical image registration.
\newblock In \emph{Proceedings of the IEEE/CVF international conference on computer vision}, pages 10600--10610, 2019.

\bibitem[Zhu and Lu(2022)]{zhu2022swin}
Y. Zhu and S. Lu.
\newblock Swin-voxelmorph: {A} symmetric unsupervised learning model for deformable medical image registration using swin transformer.
\newblock In \emph{International Conference on Medical Image Computing and Computer-Assisted Intervention}, pages 78--87. Springer, 2022.

\end{thebibliography}
}

\clearpage
\setcounter{page}{1}
\maketitlesupplementary

\section*{Appendix}

\section{Proof on distributive property of convolutions over source-target domains}
\label{App:DistrConvProof}
\subsection{Distributive property over neighbors}
Assume a feature grid \(\boldsymbol{F} \in \mathbbm{R}^{d_{in} \times H \times W \times D}\) where \(d\) is the feature dimension and \(H, W, D\) are spatial dimensions. 
A convolution over the domain \(\boldsymbol{F}\) would utilize a weight tensor \(\mathbf{W} \in \mathbbm{R}^{d_{in}\times k\times k\times k \times d_{out}}\) in a sliding-window fashion to perform a node-wise projection from domain \(\boldsymbol{F}\) to \(\boldsymbol{F'} \in \mathbbm{R}^{d_{out} \times H \times W \times D}\).
The node-wise operation at an arbitrary node \(i\) uses a neighborhood of shape \(\boldsymbol{F}_{\mathcal{N}_{i}} \in \mathbbm{R}^{d_{in}\times k\times k\times k}\) to produce an output feature \(\mathbf{f}^{i'} \in \mathbbm{R}^{d_{out}}\):
\begin{equation*}
    \mathbf{f}^{i'} = {\boldsymbol{F}_{\mathcal{N}_{i}}} \otimes \mathbf{W}
\end{equation*}

While the node-wise operation is typically implemented using tensor multiplication, a convolution only requires a weight matrix of shape \(\mathbbm{R}^{d_{in}\times d_{out}}\) to be present for each neighboring feature vector \(\mathbf{f} \in \mathbbm{R}^{d_{in}}\). 
A neighbor-wise formulation allows us to work with flattened representation of neighborhoods \(\boldsymbol{F}_{\mathcal{N}_{i}} \in \mathbbm{R}^{d_{in}\times (k \cdot k \cdot k)}\), \(\mathbf{W} \in \mathbbm{R}^{d_{in}\times (k \cdot k \cdot k) \times d_{out}}\).
This further generalizes a convolution to any arbitrary neighborhood shape outside of traditional cuboids (as long as the neighborhood shapes are consistent):
\begin{equation*}
    \mathbf{f}^{i'} = \sum_{\mathbf{f}^j \in \boldsymbol{F}_{\mathcal{N}_{i}}} \mathbf{f}^j \cdot \mathbf{W}_{[:, j]}
\end{equation*}
where \(j\) is a neighboring node to a given node \(i\).
This iterative portrayal emphasizes the distributive property of the convolution:
\begin{enumerate}
    \item First a neighbor-wise feature projection is performed using a projection matrix \(\mathbf{W}_{[:, j]} \in \mathbbm{R}^{(d_{in})\times d_{out}}\) on each element \(j\) in the \(k\times k\times k \) neighborhood
    \item Next, we perform a uniformly weighted addition of all projection vectors.
\end{enumerate}

\subsection{Distributive property over channels}
As established, each neighbor \(\mathbf{f}^j \in \boldsymbol{F}_{\mathcal{N}_{i}}\) around central node \(i\) has an independent projection matrix \(\mathbf{W}_{[:, j]} \in \mathbbm{R}^{d_{in}\times d_{out}}\).
Because of the distributive property of the dot product, the convolution operation can be further separated into a channel-wise projection:

\begin{equation*}
    \mathbf{f}^{i'} = \sum_{\mathbf{f}^j \in \boldsymbol{F}_{\mathcal{N}_{i}}} \sum_{k \in \mathbf{f}^j } k \cdot \mathbf{W}_{[k, j]}
\end{equation*}

This property comes in useful in situations where we'd like to perform the convolution operation in situations where neighborhood shapes vary across different channels.

\subsection{Source-Target domain separation}

The aforementioned properties naturally lead us to domains where concatenation is performed along feature dimensions:

Assume a feature grid \(\boldsymbol{F}^S \in \mathbbm{R}^{d_s \times H \times W \times D}\) of the source image and a feature grid \( \boldsymbol{F}^T \in \mathbbm{R}^{d_t \times H \times W \times D}\) of the target image, where \(d_s\) and \(d_t\) is the feature dimension and \(H, W, D\) are spatial dimensions. A convolution operation over the domain of concatenated source-target grids \(\boldsymbol{F} = \left[\boldsymbol{F}^S, \boldsymbol{F}^T\right] \in \mathbbm{R}^{(d_s + d_t) \times H \times W \times D}\) would have the following node-wise formulation.

\begin{equation*}
    \mathbf{f}^{i'} = \sum_{\mathbf{f}^j \in \boldsymbol{F}_{\mathcal{N}_{i}}} \mathbf{f}^j \cdot \mathbf{W}_{[:, j]} = \sum_{\left[\mathbf{f}^s, \mathbf{f}^t\right]^j \in \boldsymbol{F}_{\mathcal{N}_{i}}} \left[\mathbf{f}^s, \mathbf{f}^t\right]^j \cdot \mathbf{W}_{[:, j]}
\end{equation*}

Rearranging using distributivity across channels we get:

\begin{equation*}
    \mathbf{f}^{i'} = \sum_{\left[\mathbf{f}^s, \mathbf{f}^t\right]^j \in \boldsymbol{F}_{\mathcal{N}_{i}}} \left[\mathbf{f}^{s}\right]^j \cdot \mathbf{W}_{[0:d_s, j]} + \left[\mathbf{f}^{t}\right]^j \cdot \mathbf{W}_{[d_s:d_t, j]}
\end{equation*}
\begin{equation*}
    \mathbf{f}^{i'} = \sum_{\mathbf{f}^{j} \in \boldsymbol{F}^S_{\mathcal{N}_{i}}} \mathbf{f}^{j} \cdot \mathbf{W}_{[0:d_s, j]} + \sum_{\mathbf{f}^{j} \in \boldsymbol{F}^T_{\mathcal{N}_{i}}} \mathbf{f}^{j} \cdot \mathbf{W}_{[d_s:d_t, j]}
\end{equation*}

Since the kernel dimension is flattened, source and target neighborhoods do not need to be concatenated over spatial dimensions. This allows us to have separate neighborhood sizes for source and target domains. In fact, if both \(\boldsymbol{F}^S_{\mathcal{N}_{i}}\) and \(\boldsymbol{F}^T_{\mathcal{N}_{i}}\) are cuboid-shaped neighborhoods, we can organize both terms into two tensor-form convolution operations:

\begin{equation*}
    \mathbf{f}^{i'} = \overbrace{\boldsymbol{F}^S_{\mathcal{N}_{i}} \otimes \mathbf{W}^S}^{\text{Convolution on source domain}} + \overbrace{\boldsymbol{F}^T_{\mathcal{N}_{i}} \otimes \mathbf{W}^T}^{\text{Convolution on target domain}}
\end{equation*}

\section{Learned interpolation connection to parametric interpolation}
\label{App:Parametric_interp}

\subsection{Parametric interpolation}
A commonly adopted technique in parametric image registration involves predicting deformations at a coarse spacing and interpolating to the desired resolution via continuous mapping functions.
The domain of parametric registration offers interpolation techniques for mapping transformations with local basis functions.
These spatial parametric functions introduce highly sought-after theoretical guarantees.
Generally, these mappings are formulated in the context of a set of control points \(C\) exerting influences on the interpolation at a given point in space via local basis functions.
Particularly in the case of b-spline basis functions, the interpolation process exhibits the property of local support, implying that a small, localized change has a restricted impact and does not influence the entire domain.
 
The transformation \(\phi\) at an arbitrary point in space \(i\) with coordinates \(\mathbf{x}^i\) is the resulting interpolation of the transformation values \(\phi^c\) of its neighboring control points \(c \in \mathcal{N}(\mathbf{x^s}, I)\). 
This interpolation is weighted using basis functions \(v\), based on relative positions between the given point \(i\) and each control point \(c\):
\begin{equation}
\phi(\mathbf{x}^i)=\sum_{c \in \mathcal{N}(\mathbf{x^i}, C)} \overbrace{v\left(\mathbf{x}^c -\mathbf{x}^i\right)}^{\text{weight coefficient}} \mathbf{\phi}^c
\label{eq:B-spline}
\end{equation}
\noindent
This concept of locally weighting a transformation, based on relative location to control points, serves as a powerful heuristic for introducing local support. 
However, we argue that making the interpolation mechanism aware of image features is the key to building improved interpolation functions. 

\subsection{Cross-attention as data-driven interpolation}\label{sec:x_learned_interp}
The attention mechanism has been applied to illustrate a more general version of the convolution operation~\citep{bronstein2021geometric}. 
The convolution mechanism offers a uniformly weighted aggregation of neighbors:
\begin{equation}
    \mathbf{f}'
    = \sum_{\mathclap{j \in \mathcal{N}(\mathbf{x^i}, I)}}  \quad \overbrace{\frac{1}{\left|\mathcal{N}(\mathbf{x^i},I)\right|}}^{\text{weight coefficient}}\mathbf{f}^j \cdot W(\mathbf{x}^j - \mathbf{x}^i)
\end{equation}

Unlike the convolution's simple uniformly weighted aggregation of neighbors' responses, the attention mechanism allows a point to compute a form of learned weighted averaging based on its neighbors' features and relative positions. 

\begin{equation}
   \mathbf{f}' = \sum_{c \in \mathcal{N}(\mathbf{x^i}, C)} \overbrace{a(\mathbf{f}^c, \mathbf{f}^i, \left(\mathbf{x}^c-\mathbf{x}^i\right))}^{\text{weight coefficient}}~ \mathbf{f}^j \cdot W\!\left(\mathbf{x}^c-\mathbf{x}^i\right)
    \label{eq:Graph_att_conv}
\end{equation}
where \(\mathcal{N}(\mathbf{x^s}, I)\) is the neighborhood of control points to point \(i\). 

In the context of registration, convolutions already display desired local properties by restricting message-passing within local neighborhoods.
The attention operation extends this principle by dynamically \say{masking out} irrelevant neighboring points.

When points \(i\) and \(c\) belong to different domains (e.g., different images or resolution levels), the operation described in Eq.~(\ref{eq:Graph_att_conv}) is referred to as \textit{cross-attention}.
Here, the attention function \(a\) is constrained to be in the range \([0, 1]\) by applying a softmax operation over all neighboring control points such that \(\sum_{c \in C_{\mathcal{N}_i}}a(\cdot) = 1\).

The concept of attention as dynamically weighting neighboring points as outlined in Eq.~(\ref{eq:Graph_att_conv}) offers strong similarities to the principles of parametric registration methods outlined in Eq.~(\ref{eq:B-spline}).
Similarly to how parametric interpolation uses a preset weighting function \(v\left(\mathbf{x}^c -\mathbf{x}^i\right)\) on neighboring control points, local attention uses a learned weighting function \(a(\mathbf{f}^i, \mathbf{f}^c, \left(\mathbf{x}^c-\mathbf{x}^i\right))\).
In the local attention setting, since a given node only interacts with its spatially restricted neighborhood (and not with the entire space), a localized change does not affect the entire domain, effectively offering properties of local support. 
The benefit of the attention mechanism here is the ability to condition the weighting coefficients not only on relative coordinates but also on the learned features present in the operation.


\section{Data pre-processing}
\label{App:data_info}

Dataset, pre-processing, and label information of the CamCAN~\citep{Shafto2014TheCC,Taylor2017TheCC} and the challenging retinal optical coherence tomography (OCT) images~\cite{sutton2023developing} datasets. 
CamCAN dataset consists of \(310\) T1w and T2w MR 3D images (\(160 \times 180 \times 160\), \(1\rm{mm}^3\) isotropic resolution). 
Preprocessing includes normalization to MNI~\citep{Horn2016} space using affine registration, skull-stripping with ROBEX~\citep{Iglesias2011Robex}, and bias-field correction with SimpleITK~\citep{Lowekamp2013TheDO}. 
Automated segmentation of 138 cortical and subcortical structures (categorized into five groups for reporting) was performed using MALPEM~\citep{Ledig2015RobustWS}.
The retinal optical coherence tomography (OCT) dataset consists of \(28,741\) images accompanied by \(10\) segmentation labels per image that delineate the retinal layers.
This dataset is longitudinal, containing scans acquired at one or multiple time points for both healthy individuals and patients with age-related macular degeneration (AMD), enabling the study of temporal changes in retinal structure.
We utilize only a subset of \(1000\) images, standardized to a size of (\(32 \times 80 \times 112\) with \(1\rm{mm}^3\) isotropic resolution. 
For training, validation, and testing, we use \(80\%-10\%-10\%\) splits for both datasets.

\section{Baselines}
\label{App:Baselines}
We compare our method (GeoReg) against several conventional iterative methods and learning-based image registration models.
Regarding the iterative optimization methods, we choose from the Medical Image Registration ToolKit (MIRTK)~\citep{Schuh2014ConstructionOA}, a widely-used free-form deformation (FFD) iterative optimization method that supports multi-resolution and parametric b-spline-based registration. 
Additionally, we compare against the widely adopted symmetric diffeomorphic algorithm SyN~\citep{avants2008symmetric} from the ANTs~\citep{avants2009advanced} framework.
Our learning-based baselines are comprised of Voxelmorph~\citep{balakrishnan2019}, a single-stage CNN, LapIRN~\citep{mok2020} a multi-resolution registration CNN that aims to capture large deformations in a coarse-to-fine manner, Transmorph~\citep{chen2022transmorph} that uses a SwinTransformer-based encoder, Recursive Cascaded Networks (RCN)~\citep{zhao2019recursive} which estimates the deformation progressively using a cascading CNN architecture, the dual-stream pyramid registration network (D-PRNet)~\citep{kang2022dual} that gradually refines the multi-level predicted deformation fields in a coarse-to-fine manner via sequential warping, and Large-Kernel UNet (LK-UNet)~\cite{jia2022u} utilizes inception convolutional layers with variable (large to small) kernel sizes to allow for wider receptive field.
To ablate the contribution of the proposed interpolation mechanism (\(\delta\)) on top of our multi-resolution \(\tau\) design, we replace the proposed learned interpolation component (\(\delta\)) with bilinear feature warping. 
In the following, we denote this ablation baseline as \say{feat. warp}.

\section{Implementation details}
\label{App:implementation_details}
Our approach utilizes a lightweight dual-stream encoder design to independently extract features for the source \(S\) and target \(T\) images. 
The encoder consists of two convolutional residual blocks per layer followed by pooling layers, which allow hierarchical feature pyramid extraction at multiple scales for both source and target images.
The encoder is composed of 6 layers, each made up of two residual blocks each with \([16, 32, 32, 64, 64, 128]\) channels. Average pooling \([2\times2\times2]\) operations are applied in between each of the encoder layers.
The feature pyramid is then decoded in a coarse-to-fine fashion across the 6 resolutions.

At every resolution, there is a possibility to iteratively refine the deformation prediction by repeatedly applying \(\tau\) or skipping \(\tau\) altogether to create an interpolation-only layer.
Across our experiments, we find that local deformations are sufficiently well-modeled at coarser resolutions in the decoder, allowing our final (finest) layer to be interpolation-only layers (where \(\tau\) was not applied). 
For our synthetic deformation experiments, we applied 4 iterations of \(\tau\) on the two coarsest resolutions. 
We empirically notice that employing a larger number of iterations at the coarsest pyramid levels helps recover large transformations early in the decoding process.
This strategy also makes larger \(\tau\) iterations computationally cheap due to the fewer number of points present at the coarsest pyramid levels.
Estimating the transformation at coarser levels also reduces the registration burden of finer ones, as only smaller local deformations need to be recovered.
Neighborhood sizes of target points varied between \(3\times 3\) or \(5 \times 5\) for \(\tau\) (larger neighborhoods were used on coarser layers), while \(\delta\) always used the closest \(3\times 3\) points of the previous coarser resolution.
For further parameter details, we refer to our repository\footnote{\href{https://anonymous.4open.science/r/GeoReg-1A1D/README.md}{https://anonymous.4open.science/r/GeoReg-1A1D/README.md}}.

To calculate the loss at each resolution level, we employ normalized cross-correlation (NCC) as the dissimilarity metric.
Furthermore, we utilize bending energy~\cite{Rueckert1999NonrigidRU} as a regularize, to ensure a smooth final transformation at each resolution.
The approach is trained end-to-end using ADAM as an optimizer with a \(10^{-4}\) learning rate for a maximum of \(1000\) epochs.
Models training was carried out on an NVIDIA A40 GPU with 40GB VRAM over the course of \(3\) days.

\textbf{Memory efficient neighborhood computation.}
While the neighorhood formulation defined in the methodology is formalized using set notation, the grid structure of our data allows us to design highly memory efficient implementations of \(\tau\) and \(\delta\) layers. 
Generally, finding nearest neighbors in sparse domains is a big memory bottleneck due to having to compute \(O(N^2)\) distance calculations relative to the number of points \(N\).
However, since our convolutional encoder provides us with feature grids, we are able to use grid-unfolding operations to find the nearest neighbors in a \((k_x, k_y, k_z)\) kernel around a central node.
The \(\tau\) function's neighborhood computations are performed by first unfolding the target domain into all its possible neighbourhoods. 
Then, we can index a source node's corresponding target neighborhood by mapping the current source coordinates into the index space of the target grid and rounding to the closest integer.
Similarly, the \(\delta\) function applies a repeated interleaving operation to upsample the parent unfolded neighborhoods into the same dimensions as the children grid.
This is all implemented using standard built-in Pytorch functions that allow for efficient GPU parallelism.
Wherever possible, we make use of pointers to the original data structures for minimal memory footprints. 
We refer readers to Appendix~\ref{App:VRAM} for an overview on VRAM requirements of various registration baselines.

\newpage

\section{Training memory footprints}
\label{App:VRAM}
\begin{table}[htbp]
    \addtolength{\tabcolsep}{4pt}
    \centering
    \caption{VRAM requirements per model in GigaBytes (GB) under a batch size of 1.
    }
    \label{tab:memory}
    \setlength{\tabcolsep}{3pt}
    \begin{tabular}{l|r}
    \toprule
    Models & VRAM \\
    \midrule
    VoxelMorph & \(3.55\) GB \\
    LapIRN & \(6.67\) GB \\
    Transmorph & \(7.09\) GB \\
    D-PRNet & \(11.11\) GB \\
    RCN & \(6.21\) GB \\
    LK-UNet & \(4.18\) GB \\
    \textbf{Ours} (feat. warp) & \(6.75\) GB\\
    \textbf{Ours} (GeoReg) & \(9.08\) GB\\
    \bottomrule 
    \end{tabular}

\end{table}


\section{Architectural overview}
\label{App:architectural_overview}

\begin{figure}[h]
  \centering
  \includegraphics[width=1.9\linewidth]{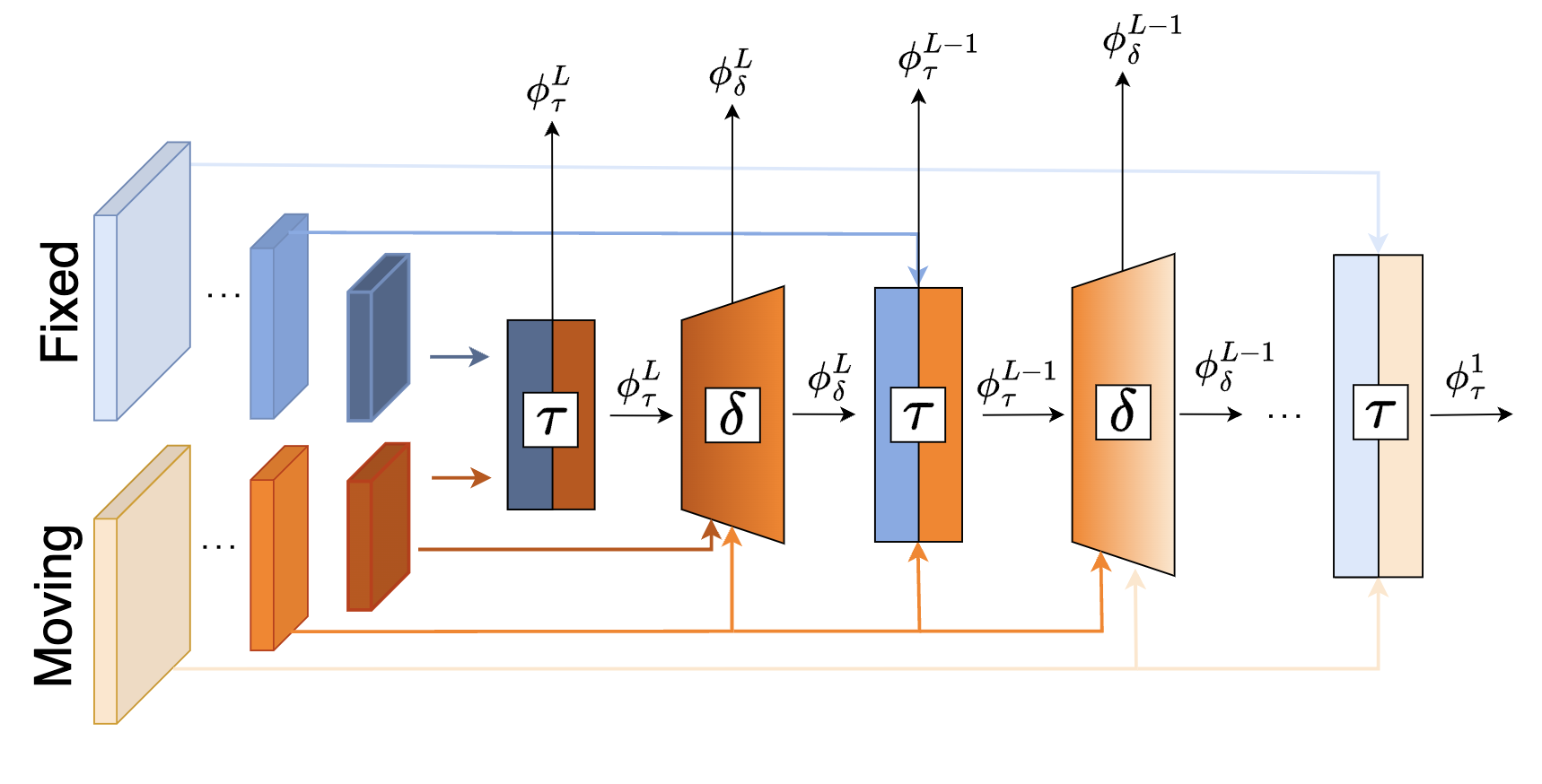}
  \caption{Architectural overview of the proposed method. A dual-stream encoders extracts features independently from source and target images. The decoder does \textbf{not} carry features across resolutions, only transformations are passed across resolutions in order to transform the source feature grids.}
\end{figure} 

\onecolumn 
\twocolumn 

\section{Pseudocode of function \texorpdfstring{\(\boldsymbol{\tau}\)}{}}
\label{App:PC_tau}

\begin{algorithm}[h]
\begin{minipage}{\linewidth}
\caption{Pseudocode of implementation of deformation function $\tau$ for a given decoder resolution}
\label{PC:tau}
    
\begin{algorithmic}
    \STATE {}
    \STATE {\underline{Function $\tau$} $(\boldsymbol{F}^S, \boldsymbol{X}^S, \boldsymbol{F}^T, \boldsymbol{X}^T, \theta^\tau)$:}
    \STATE {}
    \STATE {\textbf{Input:} Features and coordinates of source points $S$ and target points $T$. Learnable parameters $\theta^\tau$.}
    \STATE {\textbf{Output:} Deformations $\boldsymbol{U}$ for source points $S$}
    \STATE {}
    \STATE {// Initialize deformations as zeros}
    \STATE $\boldsymbol{U} \gets \boldsymbol{0}^{S\times 3}$
    \STATE {// Repeat deformation N times}
    \STATE {$n \gets 0$}
    \WHILE {$n < N$}
            \STATE {// Perform node-wise deformation on source domain}
            \FOR {$s \in S$}
            \STATE {$\mathbf{f} \gets \boldsymbol{F}^S[s]$}
            \STATE {$\phi \gets \boldsymbol{X}^S[s] + \boldsymbol{U}[s]$}
            
            \STATE {// Define target neighborhood for current source node $s$}
            \STATE {$\mathcal{N} \gets \textsc{GetNeighborhood}(\phi, \boldsymbol{X}^T)$}
            \STATE {// Perform cross-attention between a source node and its target neighborhood}
            \STATE $F^\mathcal{N} \gets \boldsymbol{F}^{T}[\mathcal{N}] + \textsc{PosEncode}(\boldsymbol{X}^{T}[\mathcal{N}] - \phi)$
            \STATE {$\mathbf{u}^T \gets \textsc{CrossAttention}(\mathbf{f}, \boldsymbol{F}^{\mathcal{N}}; \theta^\tau)$} 
            
            \STATE {// Define source neighborhood for current source node $s$}
            \STATE {$\mathcal{N} \gets \textsc{GetNeighborhood}(\phi, \boldsymbol{X}^S)$}
            \STATE {// Perform cross-attention between a source node and its source neighborhood}
            \STATE $F^\mathcal{N} \gets \boldsymbol{F}^{S}[\mathcal{N}] + \textsc{PosEncode}(\boldsymbol{X}^{S}[\mathcal{N}] - \phi)$
            \STATE {$\mathbf{u}^S \gets \textsc{CrossAttention}(\mathbf{f}, \boldsymbol{F}^{\mathcal{N}}; \theta^\tau)$} 
            \STATE {// Update total deformation estimate}
            \STATE $\mathbf{u} \gets \mathbf{u}^T + \mathbf{u}^S$ 
            \STATE $\boldsymbol{U}[s] \gets \boldsymbol{U}[s] + \mathbf{u}$
        \ENDFOR
        \STATE {$n \gets n + 1$}
    \ENDWHILE
    \STATE {return $\boldsymbol{U}$}
\end{algorithmic}
\end{minipage}
\end{algorithm}

\newpage

\section{Pseudocode of function \texorpdfstring{\(\boldsymbol{\delta}\)}{}}
\label{App:PC_delta}
\begin{algorithm}[h]
\begin{minipage}{\linewidth}
\caption{Pseudocode of implementation of deformation interpolation function $\delta$ between decoder resolution layers $r$ and $r-1$}
\label{PC:delta}
    
\begin{algorithmic}
    \STATE {}
    \STATE {\underline{Function $\delta$} $(\boldsymbol{F}^{r+1}, \boldsymbol{X}^{r+1}, \boldsymbol{U}^{r+1}, \boldsymbol{F}^{r}, \boldsymbol{X}^{r}, \theta^\delta)$:}
    \STATE {}
    \STATE {\textbf{Input:} Features $\boldsymbol{F}^{r+1}$, starting coordinates $\boldsymbol{X}^{r+1}$, and deformation $\boldsymbol{U}^{r+1}$ of source control points $S$ in layer $r+1$. Features $\boldsymbol{F}^{r}$ and coordinates $\boldsymbol{X}^{r}$ source points $S$ in layer $r$. Learnable parameters $\theta^\delta$.}
    \STATE {\textbf{Output:} Deformations $\boldsymbol{U}^{r}$ for source points $S^r$ in resolution $r$}
    
    \STATE {}
    \STATE {// Initialize deformations as zeros}
    \STATE $\boldsymbol{U}^{r} \gets \boldsymbol{0}^{S^{r}\times 3}$
    \STATE {// Perform node-wise deformation interpolation on layer $r$}
    \FOR {$s \in S^{r}$}
    \STATE {$\mathbf{f} \gets \boldsymbol{F}^{r}[s]$}
    \STATE {$\mathbf{x} \gets \boldsymbol{X}^{r}[s]$}
    \STATE {// Define neighborhood of child node $s$ prior to any deformations to parent points $S^{r+1}$.}
    \STATE {$\mathcal{N} \gets \textsc{GetNeighborhood}(\mathbf{x}, \boldsymbol{X}^{r+1})$}
    \STATE {// Compute initial deformation estimate using neighborhood mean}
    \STATE {$\mathbf{u}^{inh} \gets \textsc{BilinearInterp}(\boldsymbol{U}^{l}[\mathcal{N}_s^{S^{l}}])$}
    \STATE {$\phi \gets \mathbf{x} + \mathbf{u}^{inh}$}
    \STATE {// Perform cross-attention between child node $s$ and its neighbourhood of parent points $\mathcal{N}_s^{S^{l}}$}
    \STATE $\phi^{\mathcal{N}} \gets \boldsymbol{X}^{r+1}[\mathcal{N}] + \boldsymbol{U}^{r+1}[\mathcal{N}]$
    \STATE $\boldsymbol{F}^{\mathcal{N}} \gets \boldsymbol{F}^{r+1}[\mathcal{N}] + \textsc{PosEncode}(\phi^{\mathcal{N}} - \phi)$
    \STATE {$\mathbf{u}^{interp} \gets \textsc{CrossAttention}(\mathbf{f}, \boldsymbol{F}^{\mathcal{N}}; \theta^\delta)$} 
    \STATE $\boldsymbol{U}^{r}[s] \gets \mathbf{u}^{inh} + \mathbf{u}^{interp}$
    \ENDFOR
    \STATE {return $\boldsymbol{U}^{r}$}
\end{algorithmic}
\end{minipage}
\end{algorithm}


\onecolumn 
\twocolumn 
\section{Qualitative results on large synthetic affine transformations}
\label{App:Synth_results}
\begin{figure}[!h]
  \centering
  \includegraphics[width=1.0\textwidth]{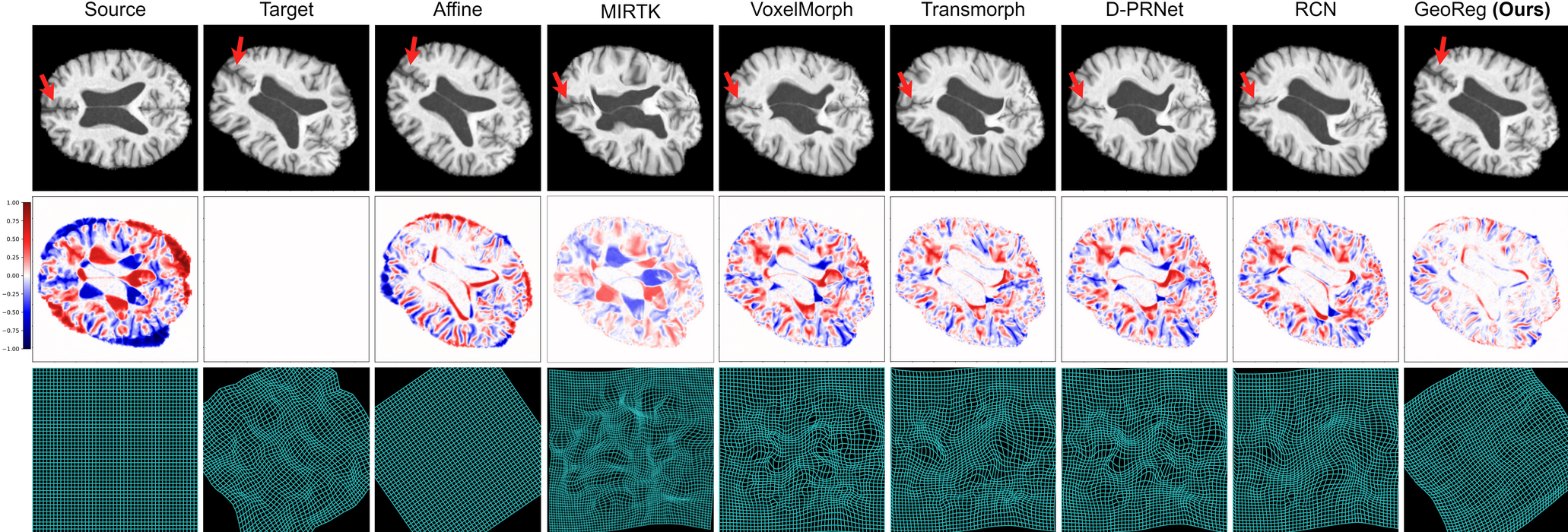}
  \caption{Qualitative results for an intra-subject \((45\degree, 0\degree, 0\degree)\) rotation experiment with added random Brownian noise deformation. Red arrows indicate the same structure across all methods. Our method is able to recover affine and deformable components despite modeling the transformation fully deformably.}
  \label{fig:affine_results}
\end{figure}

\begin{figure}[!h]
  \centering
  \includegraphics[width=1\textwidth]{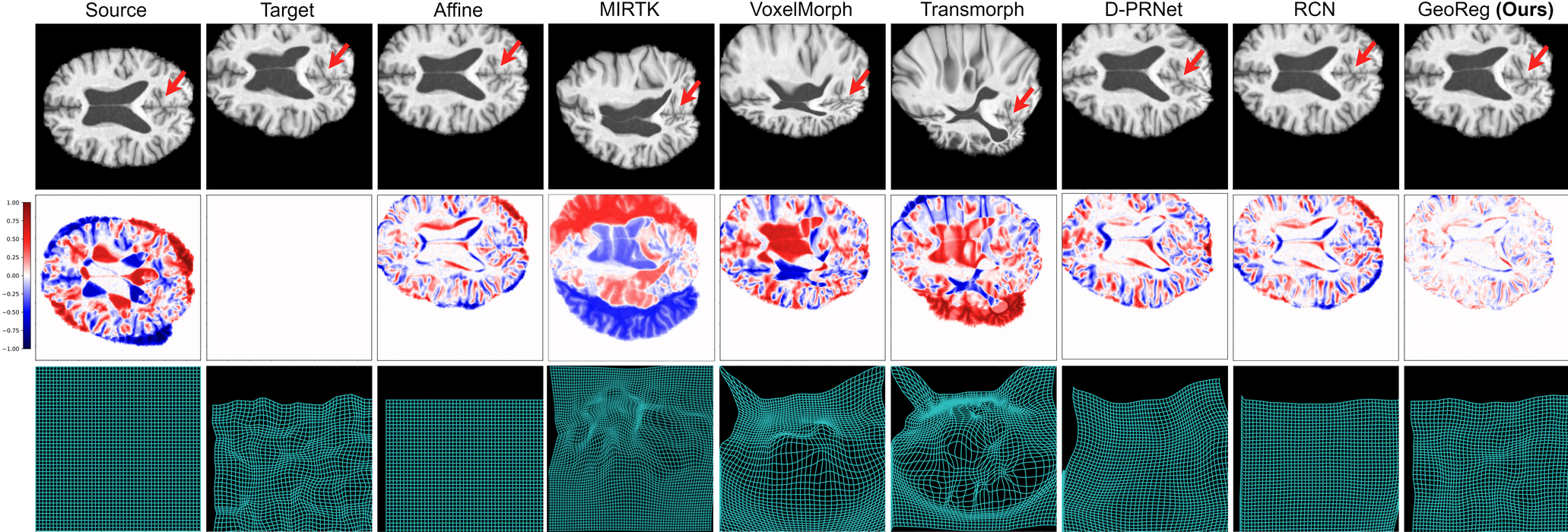}
  \caption{Qualitative results of all compared methods for an inter-subject \((25\%, 5\%, 0\%)\) of image shape translation registration experiment with added random Brownian noise deformation. Red arrows indicate the same brain structure across all registration methods. Our method is able to recover affine and deformable components despite modeling the transformation fully deformably.}
  \label{fig:tra_results}

\end{figure}

In this section we investigate varying kinds of large synthetic deformations without any form of affine registration preprocessing.
We create a dataset of intra-subject brain pairs with varying ranges of non-rigid deformations comprised of a combination of an affine and Brownian noise components.
Although the synthetic deformations may not be equivalent to real-world medical registration tasks, this experiment allows us to generate ground-truth deformations serving as a useful proof-of-concept to better evaluate recovery of large misalignments.
First, a base component of fractal Brownian deformation is applied, followed by randomly uniformly sampled rotations, scaling, and translations along each dimension (see displacement field in Figures~\ref{fig:affine_results} \&~\ref{fig:tra_results}). 

We used the obtained ground truth deformation fields to quantitatively assess a method's ability to deformably recover large misalignments.
The results reported in Table~\ref{tab:affine_results} demonstrate that our model consistently outperforms other baselines while producing the lowest amount of spatial folding. While other models struggle with large deformations, our geometric registration method is capable of fully deformably capturing the global transformation while still being able to model local deformations (see Figures~\ref{fig:affine_results} \&~\ref{fig:tra_results}). 

\begin{figure}
  \centering
  \includegraphics[width=1\textwidth]{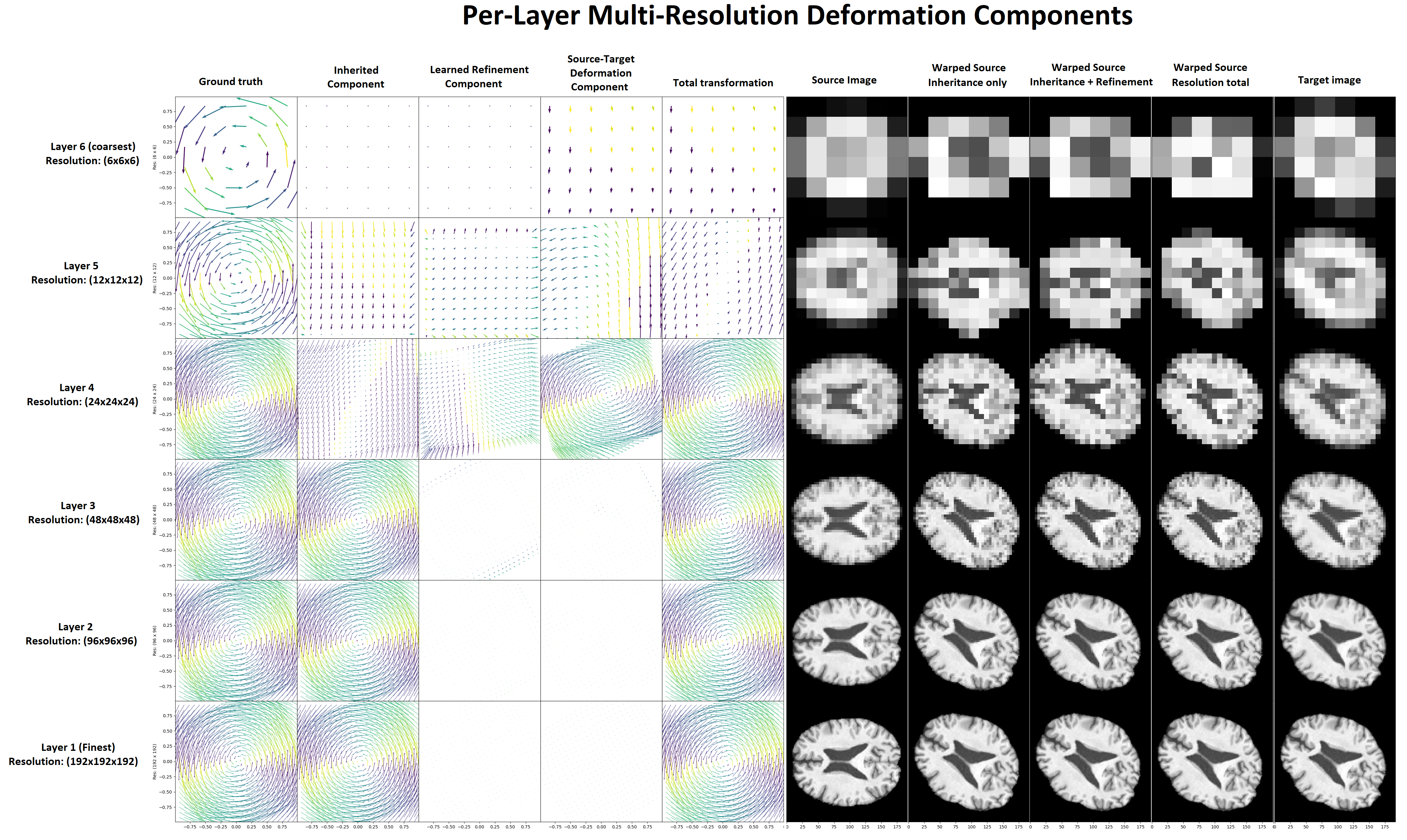}
  \caption{Per-layer deformations components as predicted by the coarse-to-fine decoding process (top-down in the figure's rows).
  Coarser layers manage to model the majority of the transformation, resulting in finer layers not having to produce meaningful deformations as their inherited transformations are already very close to ground truth transformation.}
  \label{App:fig:rot_layer_results}

\end{figure}

Figure~\ref{App:fig:rot_layer_results} displays qualitative results for intermediate transformations throughout the decoding process of a \(45\deg\) rotation augmentation.
After the 3rd resolution, the rotation transformation has been fully modeled.
Further decoder levels stop producing any meaningful further deformations.
From that point on, following decoder levels only interpolate the existing transformation.

\onecolumn 

\begin{table*}[h]
    \small
    \addtolength{\tabcolsep}{4pt}
    \centering
    \caption{Quantitative results for intra-subject deformable registration using non-rigid synthetic deformations (multi-resolution Brownian) alongside varying degrees of uniformly-sampled rotations, scalings, and translations. Lowest setting in Brownian experiment row is used as default across all other rigid rows. Experiments consist of 100 subjects, each sampled using 10 different deformations. The performance of GeoReg with bilinear feature warping instead of a learned interpolation component $\delta$ is shown under `feat. warp'.}
    \label{tab:affine_results}
    \setlength{\tabcolsep}{3pt}
    \resizebox{\columnwidth}{!}{
\begin{tabular}{lcccccccccc}
    \toprule
\multicolumn{1}{c|}{}                     & \multicolumn{1}{l|}{\# Param} & HD95 \(\downarrow\)          & AEE\(_{\phi_{GT}}\downarrow\) & \multicolumn{1}{c|}{Folding (\%) \(\downarrow\)} & HD95 \(\downarrow\)          & AEE\(_{\phi_{GT}}\downarrow\) & \multicolumn{1}{c|}{Folding (\%) \(\downarrow\)} & HD95 \(\downarrow\)           & AEE\(_{\phi_{GT}}\downarrow\) & \multicolumn{1}{c|}{Folding (\%) \(\downarrow\)} \\
    \toprule
\multicolumn{1}{c|}{\textbf{Brownian}}    & \multicolumn{1}{c|}{}         & \multicolumn{3}{c|}{Up to \(16.41\) pixels per axis (Default)}                                                              & \multicolumn{3}{c|}{Up to \(25.25\) pixels per axis}                                                               & \multicolumn{3}{c|}{Up to \(33.98\) pixels per axis}                                            \\
    \midrule
\multicolumn{1}{l|}{Affine}               & \multicolumn{1}{c|}{-}        & \(4.695 \pm 0.979\)          & \(1.813 \pm 0.316\)           & \multicolumn{1}{c|}{-}                           & \(4.821 \pm 0.992\)          & \(2.638 \pm 0.452\)           & \multicolumn{1}{c|}{-}                           & \(8.188 \pm 1.561\)           & \(2.749 \pm 0.472\)           & \multicolumn{1}{c|}{-}                           \\
\multicolumn{1}{l|}{MIRTK}                & \multicolumn{1}{c|}{-}        & \(1.940 \pm 0.170\)              & \(\mathbf{1.256 \pm 0.154}\)           & \multicolumn{1}{c|}{\(0.000 \pm 0.000\)}         & \(1.117 \pm 0.953\)          & \(1.981 \pm 0.232\)           & \multicolumn{1}{c|}{\(0.013 \pm 0.034\)}         & \(2.336 \pm 3.243\)            & \(2.635 \pm 0.239\)            & \multicolumn{1}{c|}{\(0.113 \pm 0.148\)}         \\
\multicolumn{1}{l|}{ANTs}                & \multicolumn{1}{c|}{-}        & \(2.231 \pm 0.473\)              & \(2.996 \pm 0.313\)           & \multicolumn{1}{c|}{\(0.163 \pm 0.081\)}         & \(2.781 \pm 0.815\)          & \(4.162 \pm 0.545\)           & \multicolumn{1}{c|}{\(0.282 \pm 0.065\)}         & \(4.785 \pm 1.452\)            & \(5.773 \pm 0.552\)            & \multicolumn{1}{c|}{\(0.533 \pm 0.249\)}         \\
\multicolumn{1}{l|}{LDDMM}                & \multicolumn{1}{c|}{-}        & \(1.012 \pm 0.104\)              & \(1.597 \pm 0.195\)           & \multicolumn{1}{c|}{\(0.000 \pm 0.000\)}         & \(\mathbf{1.053 \pm 0.023}\)          & \(2.443 \pm 0.238\)           & \multicolumn{1}{c|}{\(0.000 \pm 0.000\)}         & \(\mathbf{1.604 \pm 0.935}\)            & \(4.976 \pm 0.412\)            & \multicolumn{1}{c|}{\(0.195 \pm 0.015\)}         \\
\multicolumn{1}{l|}{VoxelMorph}           & \multicolumn{1}{c|}{320 k}    & \(1.656 \pm 0.159\)          & \(3.561 \pm 0.245\)           & \multicolumn{1}{c|}{\(0.003 \pm 0.002\)}         & \(3.672 \pm 0.791\)          & \(8.323 \pm 1.282\)           & \multicolumn{1}{c|}{\(0.132 \pm 0.161\)}         & \(6.199 \pm 1.910\)           & \(11.466 \pm 1.502\)          & \multicolumn{1}{c|}{\(0.249 \pm 0.203\)}         \\
\multicolumn{1}{l|}{LapIRN}              & \multicolumn{1}{c|}{924 k}    & \(- \pm -\)          & \(- \pm -\)           & \multicolumn{1}{c|}{\(- \pm -\)}         & \(- \pm -\)          & \(- \pm -\)          & \multicolumn{1}{c|}{\(- \pm -\)}         & \(- \pm -\)           & \(- \pm -\)          & \multicolumn{1}{c|}{\(- \pm -\)}         \\
\multicolumn{1}{l|}{TransMorph}           & \multicolumn{1}{c|}{46.8 M}   & \(1.010 \pm 0.025\) & \(2.247 \pm 0.049\)           & \multicolumn{1}{c|}{\(1.048 \pm 0.165\)}         & \(1.085\pm 0.088\) & \(3.468 \pm 0.109\)          & \multicolumn{1}{c|}{\(1.998 \pm 0.212\)}         & \(1.464 \pm 0.204\)           & \(3.960 \pm 0.079\)          & \multicolumn{1}{c|}{\(3.008 \pm 0.328\)}         \\
\multicolumn{1}{l|}{D-PRNet}              & \multicolumn{1}{c|}{1.2 M}    & \(1.081 \pm 0.097\)          & \(2.411 \pm 0.041\)           & \multicolumn{1}{c|}{\(0.856 \pm 0.166\)}         & \(1.424 \pm 0.175\)          & \(3.306 \pm 0.100\)          & \multicolumn{1}{c|}{\(1.645 \pm 0.202\)}         & \(2.552 \pm 0.476\)           & \(3.915 \pm 0.091\)          & \multicolumn{1}{c|}{\(2.883 \pm 0.247\)}         \\
\multicolumn{1}{l|}{RCN}                  & \multicolumn{1}{c|}{282 M}   & \(\mathbf{1.002 \pm 0.009}\)          & \(2.216 \pm 0.045\)          & \multicolumn{1}{c|}{\(1.134 \pm 0.176\)}         & \(1.087 \pm 0.074\)          & \(2.960 \pm 0.074\)           & \multicolumn{1}{c|}{\(2.249 \pm 0.213\)}          & \(2.818 \pm 0.363\)           & \(5.125 \pm 0.102\)          & \multicolumn{1}{c|}{\(1.655 \pm 0.228\)}         \\
\multicolumn{1}{l|}{FourierNet}              & \multicolumn{1}{c|}{1.1 M}    & \(1.044 \pm 0.050\)          & \(2.444 \pm 0.069\)           & \multicolumn{1}{c|}{\(0.000 \pm 0.000\)}         & \(1.350 \pm 0.123\)          & \(3.444 \pm 0.098\)          & \multicolumn{1}{c|}{\(0.001 \pm 0.001\)}         & \(1.791 \pm 0.127\)           & \(4.669 \pm 0.137\)          & \multicolumn{1}{c|}{\(0.002 \pm 0.003\)}         \\
\multicolumn{1}{l|}{\textbf{Ours} (feat. warp)}                 & \multicolumn{1}{c|}{1.5 M}    & \(2.621 \pm 0.502\)          & \(1.637 \pm 0.161\)  & \multicolumn{1}{c|}{\(0.000 \pm 0.000\)}         & \(3.923 \pm 0.594\)          & \(2.638 \pm 0.452\)  & \multicolumn{1}{c|}{\(0.000 \pm 0.000\)}         & \(3.939 \pm 1.150\) & \(2.801 \pm 0.280\)  & \multicolumn{1}{c|}{\(0.000 \pm 0.000\)}        \\
\multicolumn{1}{l|}{\textbf{Ours} (GeoReg)}                 & \multicolumn{1}{c|}{1.7 M}    & \(1.347 \pm 0.397\)          & \(1.328 \pm 0.152\)  & \multicolumn{1}{c|}{\(0.000 \pm 0.000\)}         & \(1.763 \pm 0.421\)          & \(\mathbf{1.831 \pm 0.193}\)  & \multicolumn{1}{c|}{\(0.000 \pm 0.000\)}         & \(2.460 \pm 0.591\) & \(\mathbf{2.580 \pm 0.303}\)  & \multicolumn{1}{c|}{\(0.000 \pm 0.000\)}        \\

    \toprule
\multicolumn{1}{c|}{\textbf{Rotation + Brownian}}    & \multicolumn{1}{c|}{}         & \multicolumn{3}{c|}{\(\pm 11.25\degree\) per axis}                                                              & \multicolumn{3}{c|}{\(\pm 22.5\degree\) per axis}                                                               & \multicolumn{3}{c|}{\(\pm 45.0\degree\) per axis}                                            \\
    \midrule
\multicolumn{1}{l|}{Affine}               & \multicolumn{1}{c|}{-}        & \(4.573 \pm 0.291\)          & \(3.686 \pm 0.098\)           & \multicolumn{1}{c|}{-}                           & \(4.599 \pm 0.331\)          & \(3.682 \pm 0.109\)           & \multicolumn{1}{c|}{-}                           & \(4.600 \pm 0.369\)           & \(3.809 \pm 0.110\)           & \multicolumn{1}{c|}{-}                           \\
\multicolumn{1}{l|}{MIRTK}                & \multicolumn{1}{c|}{-}        & \(\mathbf{1.041 \pm 0.124}\)              & \(3.685 \pm 3.029\)           & \multicolumn{1}{c|}{\(0.031 \pm 0.082\)}         & \(3.515 \pm 4.905\)          & \(9.78 \pm 9.292\)           & \multicolumn{1}{c|}{\(0.265 \pm 0.392\)}         & \(6.839 \pm 8.975\)            & \(8.851 \pm 7.453\)            & \multicolumn{1}{c|}{\(0.160 \pm 0.204\)}         \\
\multicolumn{1}{l|}{ANTs}              & \multicolumn{1}{c|}{-}    & \(3.870 \pm 1.491\)          & \(7.011 \pm 3.616\)           & \multicolumn{1}{c|}{\(0.188 \pm 0.077\)}         & \(8.813 \pm 2.152\)          & \(18.571 \pm 3.813\)          & \multicolumn{1}{c|}{\(0.215 \pm 0.088\)}         & \(11.617 \pm 5.625\)           & \(30.327 \pm 14.763\)          & \multicolumn{1}{c|}{\(0.485 \pm 0.368\)}         \\
\multicolumn{1}{l|}{LDDMM}              & \multicolumn{1}{c|}{-}    & \(1.150 \pm 0.012\)          & \(5.765 \pm 3.619\)           & \multicolumn{1}{c|}{\(0.000 \pm 0.000\)}         & \(\mathbf{1.041 \pm 0.124}\)          & \(13.301 \pm 6.466\)          & \multicolumn{1}{c|}{\(0.000 \pm 0.000\)}         & \(5.843 \pm 6.615\)           & \(34.077 \pm 7.573\)          & \multicolumn{1}{c|}{\(0.013 \pm 0.049\)}         \\
\multicolumn{1}{l|}{VoxelMorph}           & \multicolumn{1}{c|}{320 k}    & \(1.816 \pm 0.298\)          & \(6.673 \pm 1.054\)           & \multicolumn{1}{c|}{\(0.034 \pm 0.020\)}         & \(3.474 \pm 0.713\)          & \(13.591 \pm 2.318\)          & \multicolumn{1}{c|}{\(0.097 \pm 0.041\)}         & \(8.997 \pm 2.353\)           & \(27.090 \pm 5.130\)          & \multicolumn{1}{c|}{\(0.292 \pm 0.077\)}         \\
\multicolumn{1}{l|}{LapIRN}              & \multicolumn{1}{c|}{924 k}    & \(- \pm -\)          & \(- \pm -\)           & \multicolumn{1}{c|}{\(- \pm -\)}         & \(- \pm -\)          & \(- \pm -\)          & \multicolumn{1}{c|}{\(- \pm -\)}         & \(- \pm -\)           & \(- \pm -\)          & \multicolumn{1}{c|}{\(- \pm -\)}         \\
\multicolumn{1}{l|}{TransMorph}           & \multicolumn{1}{c|}{46.8 M}   & \(1.057 \pm 0.073\) & \(5.087 \pm 0.775\)           & \multicolumn{1}{c|}{\(3.030 \pm 0.514\)}         & \(1.420 \pm 0.385\) & \(11.334 \pm 2.999\)          & \multicolumn{1}{c|}{\(3.560 \pm 0.414\)}         & \(5.747 \pm 2.289\)           & \(26.394 \pm 5.301\)          & \multicolumn{1}{c|}{\(4.012 \pm 0.329\)}         \\
\multicolumn{1}{l|}{D-PRNet}              & \multicolumn{1}{c|}{1.2 M}    & \(1.557 \pm 0.367\)          & \(7.002 \pm 1.202\)           & \multicolumn{1}{c|}{\(1.422 \pm 0.188\)}         & \(3.580 \pm 1.082\)          & \(14.058 \pm 2.896\)          & \multicolumn{1}{c|}{\(1.629 \pm 0.265\)}         & \(9.200 \pm 2.444\)           & \(28.278 \pm 5.769\)          & \multicolumn{1}{c|}{\(2.192 \pm 0.313\)}         \\
\multicolumn{1}{l|}{RCN}                  & \multicolumn{1}{c|}{282 M}    & \(1.364 \pm 0.130\)          & \(4.262 \pm 0.518\)           & \multicolumn{1}{c|}{\(3.640 \pm 0.698\)}         & \(1.902 \pm 0.218\)          & \(11.082 \pm 2.042\)          & \multicolumn{1}{c|}{\(3.945 \pm 0.558\)}         & \(4.951 \pm 1.777\)           & \(26.537 \pm 6.120\)          & \multicolumn{1}{c|}{\(4.029 \pm 0.505\)}         \\
\multicolumn{1}{l|}{FourierNet}              & \multicolumn{1}{c|}{1.1 M}    & \(2.224 \pm 1.205\)          & \(8.804 \pm 4.194\)      & \multicolumn{1}{c|}{\(0.000 \pm 0.000\)}         & \(4.875 \pm 3.666\)          & \(16.706 \pm 8.034\)          & \multicolumn{1}{c|}{\(0.000 \pm 0.000\)}         & \(14.253 \pm 5.661\)           & \(36.493 \pm 13.914\)          & \multicolumn{1}{c|}{\(0.007 \pm 0.016\)}         \\
\multicolumn{1}{l|}{\textbf{Ours} (feat. warp)}                 & \multicolumn{1}{c|}{1.5 M}    & \(2.068 \pm 0.484\)          & \(1.585 \pm 0.312\)  & \multicolumn{1}{c|}{\(0.000 \pm 0.000\)}         & \(2.620 \pm 1.358\)          & \(1.989 \pm 0.753\)  & \multicolumn{1}{c|}{\(0.000 \pm 0.000\)}         & \(2.818 \pm 0.546\) & \(2.477 \pm 0.928\)  & \multicolumn{1}{c|}{\(0.000 \pm 0.000\)}        \\
\multicolumn{1}{l|}{\textbf{Ours} (GeoReg)}                 & \multicolumn{1}{c|}{1.7 M}    & \(1.520 \pm 0.332\)          & \(\mathbf{1.511 \pm 0.260}\)  & \multicolumn{1}{c|}{\(0.000 \pm 0.000\)}         & \(1.630 \pm 0.415\)          & \(\mathbf{1.604 \pm 0.363}\)  & \multicolumn{1}{c|}{\(0.000 \pm 0.000\)}         & \(\mathbf{2.054 \pm 0.385}\) & \(\mathbf{1.951 \pm 0.598}\)  & \multicolumn{1}{c|}{\(0.026 \pm 0.145\)}        \\

    \toprule
\multicolumn{1}{c|}{\textbf{Scaling + Brownian}}     & \multicolumn{1}{c|}{}         & \multicolumn{3}{c|}{\(\pm 10\%\) of image size per axis}                                                        & \multicolumn{3}{c|}{\(\pm 30\%\) of image size per axis}                                                        & \multicolumn{3}{c|}{\(\pm 50\%\) of image size per axis}                                     \\
    \midrule
\multicolumn{1}{l|}{Affine}               & \multicolumn{1}{c|}{-}        & \(4.529 \pm 0.368\)          & \(3.685 \pm 0.101\)           & \multicolumn{1}{c|}{-}                           & \(4.891 \pm 0.431\)          & \(3.687 \pm 0.113\)           & \multicolumn{1}{c|}{-}                           & \(5.138 \pm 0.515\)           & \(3.749 \pm 0.116\)           & \multicolumn{1}{c|}{-}                           \\
\multicolumn{1}{l|}{MIRTK}                & \multicolumn{1}{c|}{-}        & \(1.052 \pm 0.127\)              & \(1.462 \pm 0.348\)           & \multicolumn{1}{c|}{\(0.039 \pm 0.002\)}         & \(4.545 \pm 1.598\)          & \(1.554 \pm 0.284\)           & \multicolumn{1}{c|}{\(0.398 \pm 0.699\)}         & \(9.780 \pm 11.523\)           & \(13.425 \pm 11.322\)           & \multicolumn{1}{c|}{\(0.581 \pm 0.838\)}         \\
\multicolumn{1}{l|}{ANTs}              & \multicolumn{1}{c|}{-}    & \(3.124 \pm 0.584\)          & \(0.584 \pm 1.113\)           & \multicolumn{1}{c|}{\(0.202 \pm 0.117\)}         & \(9.343 \pm 4.464\)          & \(14.023 \pm 4.234\)          & \multicolumn{1}{c|}{\(0.242 \pm 0.105\)}         & \(14.641 \pm 4.983\)           & \(20.269 \pm 4.888\)          & \multicolumn{1}{c|}{\(0.191 \pm 0.097\)}         \\
\multicolumn{1}{l|}{LDDMM}              & \multicolumn{1}{c|}{-}    & \(1.563 \pm 0.342\)          & \(2.497 \pm 0.700\)           & \multicolumn{1}{c|}{\(0.000 \pm 0.000\)}         & \(1.902 \pm 0.235\)          & \(4.765 \pm 2.273\)          & \multicolumn{1}{c|}{\(0.000 \pm 0.000\)}         & \(1.883 \pm 0.166\)           & \(8.979 \pm 4.21\)          & \multicolumn{1}{c|}{\(0.000 \pm 0.000\)}         \\
\multicolumn{1}{l|}{VoxelMorph}           & \multicolumn{1}{c|}{320 k}    & \(1.706 \pm 0.167\)          & \(3.542 \pm 0.242\)           & \multicolumn{1}{c|}{\(0.002 \pm 0.002\)}         & \(3.389 \pm 0.642\)          & \(7.074 \pm 1.045\)           & \multicolumn{1}{c|}{\(0.122 \pm 0.130\)}         & \(7.592 \pm 2.113\)           & \(12.287 \pm 1.798\)          & \multicolumn{1}{c|}{\(0.228 \pm 0.116\)}         \\
\multicolumn{1}{l|}{LapIRN}              & \multicolumn{1}{c|}{924 k}    & \(- \pm -\)          & \(- \pm -\)           & \multicolumn{1}{c|}{\(- \pm -\)}         & \(- \pm -\)          & \(- \pm -\)          & \multicolumn{1}{c|}{\(- \pm -\)}         & \(- \pm -\)           & \(- \pm -\)          & \multicolumn{1}{c|}{\(- \pm -\)}         \\
\multicolumn{1}{l|}{Transmorph}           & \multicolumn{1}{c|}{46.8 M}   & \(1.074 \pm 0.079\)          & \(3.308 \pm 0.214\)           & \multicolumn{1}{c|}{\(1.536 \pm 0.213\)}         & \(1.370 \pm 0.363\)          & \(6.384 \pm 0.635\)           & \multicolumn{1}{c|}{\(3.753 \pm 0.612\)}         & \(3.154 \pm 1.531\)           & \(10.307 \pm 1.705\)          & \multicolumn{1}{c|}{\(5.086 \pm 0.621\)}         \\
\multicolumn{1}{l|}{D-PRNet}              & \multicolumn{1}{c|}{1.2 M}    & \(1.250 \pm 0.137\)          & \(3.598 \pm 0.270\)           & \multicolumn{1}{c|}{\(1.388 \pm 0.318\)}         & \(2.225 \pm 0.347\)          & \(7.505 \pm 1.028\)           & \multicolumn{1}{c|}{\(3.200 \pm 0.445\)}         & \(5.391 \pm 2.160\)           & \(11.986 \pm 2.402\)          & \multicolumn{1}{c|}{\(3.910 \pm 0.465\)}         \\
\multicolumn{1}{l|}{RCN}                  & \multicolumn{1}{c|}{282 M}    & \(1.337 \pm 0.188\)          & \(3.307 \pm 0.181\)           & \multicolumn{1}{c|}{\(1.600 \pm 0.370\)}         & \(2.593 \pm 0.252\)          & \(5.396 \pm 0.586\)           & \multicolumn{1}{c|}{\(4.642 \pm 0.806\)}         & \(3.785 \pm 0.661\)           & \(7.834 \pm 1.276\)           & \multicolumn{1}{c|}{\(5.644 \pm 0.687\)}         \\
\multicolumn{1}{l|}{FourierNet}              & \multicolumn{1}{c|}{1.1 M}    & \(1.307 \pm 0.302\)          & \(3.831 \pm 0.601\)           & \multicolumn{1}{c|}{\(0.000 \pm 0.000\)}         & \(5.068 \pm 4.076\)          & \(9.190 \pm 3.395\)          & \multicolumn{1}{c|}{\(0.062 \pm 0.069\)}         & \(10.102 \pm 4.565\)           & \(20.638 \pm 4.237\)          & \multicolumn{1}{c|}{\(0.114 \pm 0.123\)}         \\
\multicolumn{1}{l|}{\textbf{Ours} (feat. warp)}                 & \multicolumn{1}{c|}{1.5 M}    & \(1.910 \pm 0.355\) & \(1.490 \pm 0.277\)  & \multicolumn{1}{c|}{\(0.000 \pm 0.000\)}         & \(2.566 \pm 0.617\) & \(2.073 \pm 0.925\)  & \multicolumn{1}{c|}{\(0.000 \pm 0.000\)}         & \(2.961 \pm 0.796\)  & \(2.400 \pm 1.290\)  & \multicolumn{1}{c|}{\(0.000 \pm 0.000\)}         \\
\multicolumn{1}{l|}{\textbf{Ours} (GeoReg)}                 & \multicolumn{1}{c|}{1.7 M}    & \(\mathbf{1.040 \pm 0.122}\) & \(\mathbf{1.375 \pm 0.357}\)  & \multicolumn{1}{c|}{\(0.000 \pm 0.000\)}         & \(\mathbf{1.274 \pm 0.312}\) & \(\mathbf{1.714 \pm 0.904}\)  & \multicolumn{1}{c|}{\(0.000 \pm 0.000\)}         & \(\mathbf{1.486 \pm 0.523}\)  & \(\mathbf{2.234 \pm 1.586}\)  & \multicolumn{1}{c|}{\(0.000 \pm 0.000\)}         \\
    \toprule
\multicolumn{1}{c|}{\textbf{Translation + Brownian}} & \multicolumn{1}{c|}{}         & \multicolumn{3}{c|}{\(\pm 10\%\) of image size per axis}                                                        & \multicolumn{3}{c|}{\(\pm 30\%\) of image size per axis}                                                        & \multicolumn{3}{c|}{\(\pm 50\%\) of image size per axis}                                     \\
    \midrule
\multicolumn{1}{l|}{Affine}               & \multicolumn{1}{c|}{-}        & \(4.791 \pm 1.106\)          & \(2.092 \pm 0.362\)           & \multicolumn{1}{c|}{-}                           & \(4.683 \pm 1.241\)          & \(2.076 \pm 0.386\)           & \multicolumn{1}{c|}{-}                           & \(4.768 \pm 1.088\)           & \(2.025 \pm 0.497\)           & \multicolumn{1}{c|}{-}                           \\
\multicolumn{1}{l|}{MIRTK}                & \multicolumn{1}{c|}{-}         & \(2.217 \pm 0.256\)          & \(1.583 \pm 0.418\)            & \multicolumn{1}{c|}{\(0.030 \pm 0.172\)}        & \(15.738 \pm 9.906\)         & \(15.912 \pm 7.513\)           & \multicolumn{1}{c|}{\(0.920 \pm 0.549\)}         & \(31.954 \pm 18.177\)         & \(32.458 \pm 18.79\)           & \multicolumn{1}{c|}{\(0.557 \pm 0.413\)}         \\
\multicolumn{1}{l|}{ANTs}              & \multicolumn{1}{c|}{-}    & \(2.641 \pm 1.983\)          & \(4.269 \pm 1.888\)           & \multicolumn{1}{c|}{\(0.191 \pm 0.097\)}         & \(20.516 \pm 6.970\)          & \(21.84 \pm 7.159\)          & \multicolumn{1}{c|}{\(0.206 \pm 0.139\)}         & \(39.502 \pm 13.644\)           & \(42.077 \pm 14.147\)          & \multicolumn{1}{c|}{\(0.139 \pm 0.053\)}         \\
\multicolumn{1}{l|}{LDDMM}              & \multicolumn{1}{c|}{-}    & \(1.962 \pm 0.310\)          & \(3.195 \pm 0.806\)           & \multicolumn{1}{c|}{\(0.000 \pm 0.000\)}         & \(15.476 \pm 10.564\)          & \(14.518 \pm 6.879\)          & \multicolumn{1}{c|}{\(0.000 \pm 0.000\)}         & \(31.702 \pm 17.558\)           & \(30.984 \pm 13.721\)          & \multicolumn{1}{c|}{\(0.549 \pm 1.403\)}         \\
\multicolumn{1}{l|}{VoxelMorph}           & \multicolumn{1}{c|}{320 k}    & \(3.468 \pm 0.529\)          & \(5.436 \pm 0.495\)           & \multicolumn{1}{c|}{\(0.033 \pm 0.034\)}         & \(18.075 \pm 2.677\)         & \(16.027 \pm 1.977\)          & \multicolumn{1}{c|}{\(0.628 \pm 0.194\)}         & \(31.645 \pm 4.145\)          & \(26.826 \pm 3.932\)          & \multicolumn{1}{c|}{\(1.479 \pm 0.395\)}         \\
\multicolumn{1}{l|}{LapIRN}              & \multicolumn{1}{c|}{924 k}    & \(- \pm -\)          & \(- \pm -\)           & \multicolumn{1}{c|}{\(- \pm -\)}         & \(- \pm -\)          & \(- \pm -\)          & \multicolumn{1}{c|}{\(- \pm -\)}         & \(- \pm -\)           & \(- \pm -\)          & \multicolumn{1}{c|}{\(- \pm -\)}         \\
\multicolumn{1}{l|}{TransMorph}           & \multicolumn{1}{c|}{46.8 M}   & \(1.641 \pm 0.385\)          & \(6.065 \pm 0.634\)           & \multicolumn{1}{c|}{\(2.601 \pm 0.421\)}         & \(17.865 \pm 4.339\)         & \(20.212 \pm 3.201\)          & \multicolumn{1}{c|}{\(4.378 \pm 0.275\)}         & \(40.148 \pm 7.138\)          & \(37.221 \pm 5.569\)          & \multicolumn{1}{c|}{\(5.920 \pm 0.147\)}         \\
\multicolumn{1}{l|}{D-PRNet}              & \multicolumn{1}{c|}{1.2 M}    & \(3.720 \pm 0.879\)         & \(6.045 \pm 0.751\)         & \multicolumn{1}{c|}{\(2.631 \pm 0.483\)}         & \(5.477 \pm 0.692\)          & \(12.834 \pm 1.757\)          & \multicolumn{1}{c|}{\(5.556 \pm 0.428\)}         & \(6.669 \pm 0.833\)           & \(17.522 \pm 2.954\)          & \multicolumn{1}{c|}{\(6.636 \pm 0.509\)}         \\
\multicolumn{1}{l|}{RCN}                  & \multicolumn{1}{c|}{282 M}    & \(2.217 \pm 0.193\)          & \(3.559 \pm 0.163\)           & \multicolumn{1}{c|}{\(4.465 \pm 1.071\)}         & \(4.632 \pm 0.352\)          & \(6.427 \pm 0.608\)           & \multicolumn{1}{c|}{\(6.491 \pm 0.883\)}         & \(5.123 \pm 0.430\)           & \(10.281 \pm 1.381\)          & \multicolumn{1}{c|}{\(7.166 \pm 0.565\)}         \\
\multicolumn{1}{l|}{FourierNet}              & \multicolumn{1}{c|}{1.1 M}    & \(1.790 \pm 0.139\)          & \(2.920 \pm 0.062\)           & \multicolumn{1}{c|}{\(0.019 \pm 0.029\)}         & \(4.762 \pm 0.325\)          & \(3.737 \pm 0.116\)          & \multicolumn{1}{c|}{\(0.146 \pm 0.109\)}         & \(4.664 \pm 0.470\)           & \(4.032 \pm 0.177\)          & \multicolumn{1}{c|}{\(0.087 \pm 0.074\)}         \\
\multicolumn{1}{l|}{\textbf{Ours} (feat. warp)}                 & \multicolumn{1}{c|}{1.5 M}    & \(2.193 \pm 0.334\) & \(1.474 \pm 0.156\)  & \multicolumn{1}{c|}{\(0.000 \pm 0.000\)}         & \(3.397 \pm 0.446\) & \(1.949 \pm 0.189\)  & \multicolumn{1}{c|}{\(0.000 \pm 0.000\)}         & \(4.605 \pm 2.827\)  & \(3.279 \pm 2.685\)  & \multicolumn{1}{c|}{\(0.000 \pm 0.000\)}    \\
\multicolumn{1}{l|}{\textbf{Ours} (GeoReg)}                 & \multicolumn{1}{c|}{1.7 M}    & \(\mathbf{1.293 \pm 0.308}\) & \(\mathbf{1.288 \pm 0.161}\)  & \multicolumn{1}{c|}{\(0.000 \pm 0.000\)}         & \(\mathbf{1.603 \pm 0.329}\) & \(\mathbf{1.434 \pm 0.205}\)  & \multicolumn{1}{c|}{\(0.000 \pm 0.000\)}         & \(\mathbf{2.260 \pm 0.358}\)  & \(\mathbf{1.760 \pm 0.295}\)  & \multicolumn{1}{c|}{\(0.000 \pm 0.000\)}   \\

    \toprule
\multicolumn{1}{c|}{\textbf{Affine + Brownian}} & \multicolumn{1}{c|}{}         & \multicolumn{3}{c|}{\(\pm 11.25\degree\) Rot., \(\pm 10\%\) Scale, \(\pm 10\%\) Transl.}                                                        & \multicolumn{3}{c|}{\(\pm 22.5\degree\) Rot., \(\pm 30\%\) Scale, \(\pm 30\%\) Transl.}                                                        & \multicolumn{3}{c|}{\(\pm 45.0\degree\) Rot., \(\pm 50\%\) Scale, \(\pm 50\%\) Transl.}                                     \\
    \midrule
\multicolumn{1}{l|}{Affine}               & \multicolumn{1}{c|}{-}        & \(4.646 \pm 1.339\)          & \(3.106 \pm 1.009\)           & \multicolumn{1}{c|}{-}                           & \(4.741 \pm 1.381\)          & \(4.612 \pm 0.294\)           & \multicolumn{1}{c|}{-}                           & \(4.931 \pm 2.165\)           & \(6.388 \pm 3.766\)           & \multicolumn{1}{c|}{-}                           \\
\multicolumn{1}{l|}{MIRTK}                & \multicolumn{1}{c|}{-}         & \(1.182 \pm 0.548\)          & \(4.348 \pm 2.152\)            & \multicolumn{1}{c|}{\(0.039 \pm 0.102\)}        & \(14.953 \pm 13.197\)         & \(17.389 \pm 10.896\)           & \multicolumn{1}{c|}{\(0.442 \pm 0.456\)}         & \(54.075 \pm 13.242\)         & \(45.730 \pm 13.964\)           & \multicolumn{1}{c|}{\(1.188 \pm 0.882\)}         \\
\multicolumn{1}{l|}{ANTs}              & \multicolumn{1}{c|}{-}    & \(8.933 \pm 2.450\)          & \(11.500 \pm 3.646\)           & \multicolumn{1}{c|}{\(0.170 \pm 0.071\)}         & \(22.646 \pm 6.038\)          & \(27.093 \pm 5.549\)          & \multicolumn{1}{c|}{\(0.159 \pm 0.059\)}         & \(45.812 \pm 13.148\)           & \(53.287 \pm 7.424\)          & \multicolumn{1}{c|}{\(0.523 \pm 0.535\)}         \\
\multicolumn{1}{l|}{LDDMM}              & \multicolumn{1}{c|}{-}    & \(1.885 \pm 2.655\)          & \(8.062 \pm 3.118\)           & \multicolumn{1}{c|}{\(0.000 \pm 0.000\)}         & \(16.810 \pm 10.331\)          & \(21.628 \pm 7.729\)          & \multicolumn{1}{c|}{\(0.064 \pm 0.193\)}         & \(39.410 \pm 16.364\)           & \(51.512 \pm 12.734\)          & \multicolumn{1}{c|}{\(0.061 \pm 0.139\)}         \\
\multicolumn{1}{l|}{VoxelMorph}           & \multicolumn{1}{c|}{320 k}    & \(3.144 \pm 2.181\)          & \(4.464 \pm 1.065\)           & \multicolumn{1}{c|}{\(0.034 \pm 0.061\)}         & \(23.367 \pm 12.870\)         & \(9.264 \pm 1.842\)          & \multicolumn{1}{c|}{\(0.983 \pm 0.714\)}         & \(34.688 \pm 15.383\)          & \(13.938 \pm 5.523\)          & \multicolumn{1}{c|}{\(2.237 \pm 1.810\)}         \\
\multicolumn{1}{l|}{LapIRN}              & \multicolumn{1}{c|}{924 k}    & \(- \pm -\)          & \(- \pm -\)           & \multicolumn{1}{c|}{\(- \pm -\)}         & \(- \pm -\)          & \(- \pm -\)          & \multicolumn{1}{c|}{\(- \pm -\)}         & \(- \pm -\)           & \(- \pm -\)          & \multicolumn{1}{c|}{\(- \pm -\)}         \\
\multicolumn{1}{l|}{TransMorph}           & \multicolumn{1}{c|}{46.8 M}   & \(1.215 \pm 0.459\)          & \(6.221 \pm 1.813\)           & \multicolumn{1}{c|}{\(4.738 \pm 0.926\)}         & \(3.489 \pm 3.174\)         & \(15.329 \pm 6.015\)          & \multicolumn{1}{c|}{\(20.046 \pm 16.595\)}         & \(39.186 \pm 15.634\)          & \(30.938 \pm 11.596\)          & \multicolumn{1}{c|}{\(7.950 \pm 1.028\)}         \\
\multicolumn{1}{l|}{D-PRNet}              & \multicolumn{1}{c|}{1.2 M}    & \(4.081 \pm 1.298\)         & \(9.402 \pm 0.939\)         & \multicolumn{1}{c|}{\(0.258 \pm 0.262\)}         & \(11.621 \pm 2.591\)          & \(19.445 \pm 3.051\)          & \multicolumn{1}{c|}{\(6.128 \pm 0.506\)}         & \(27.491 \pm 5.682\)           & \(41.544 \pm 5.765\)          & \multicolumn{1}{c|}{\(6.709 \pm 0.318\)}         \\
\multicolumn{1}{l|}{RCN}                  & \multicolumn{1}{c|}{282 M}    & \(\mathbf{1.023 \pm 0.034}\)          & \(3.749 \pm 0.297\)           & \multicolumn{1}{c|}{\(5.368 \pm 0.355\)}         & \(\mathbf{2.006 \pm 0.160}\)          & \(9.279 \pm 0.932\)           & \multicolumn{1}{c|}{\(7.120 \pm 0.305\)}         & \(5.014 \pm 1.692\)           & \(23.953 \pm 4.785\)          & \multicolumn{1}{c|}{\(7.563 \pm 0.510\)}         \\
\multicolumn{1}{l|}{FourierNet}              & \multicolumn{1}{c|}{1.1 M}    & \(1.469 \pm 0.096\)          & \(2.825 \pm 0.070\)           & \multicolumn{1}{c|}{\(0.001 \pm 0.001\)}         & \(2.496 \pm 0.203\)          & \(3.561 \pm 0.115\)          & \multicolumn{1}{c|}{\(0.003 \pm 0.005\)}         & \(4.221 \pm 0.234\)           & \(\mathbf{4.005 \pm 0.100}\)          & \multicolumn{1}{c|}{\(0.004 \pm 0.006\)}         \\
\multicolumn{1}{l|}{\textbf{Ours} (feat. warp)}                 & \multicolumn{1}{c|}{1.5 M}    & \(2.384 \pm 0.491\) & \(1.854 \pm 0.369\)  & \multicolumn{1}{c|}{\(0.000 \pm 0.000\)}         & \(2.982 \pm 1.089\) & \(3.031 \pm 1.016\)  & \multicolumn{1}{c|}{\(0.000 \pm 0.000\)}         & \(\mathbf{4.109 \pm 2.371}\)  & \(6.436 \pm 3.811\)  & \multicolumn{1}{c|}{\(0.001 \pm 0.001\)}    \\
\multicolumn{1}{l|}{\textbf{Ours} (GeoReg)}                 & \multicolumn{1}{c|}{1.7 M}    & \(1.880\pm 0.431\) & \(\mathbf{1.614 \pm 0.239}\)  & \multicolumn{1}{c|}{\(0.000 \pm 0.000\)}         & \(3.567 \pm 0.960\) & \(\mathbf{2.576 \pm 0.703}\)  & \multicolumn{1}{c|}{\(0.000 \pm 0.000\)}         & \(6.275 \pm 3.026\)  & \(4.437 \pm 2.774\)  & \multicolumn{1}{c|}{\(0.063 \pm 0.347\)}   \\
    \bottomrule
\end{tabular}
 }
\end{table*}

\twocolumn 
\onecolumn

\newpage
\section{Qualitative results}
\label{App:qual_def}

\begin{figure}[b]
  \centering
  \includegraphics[width=1\textwidth]{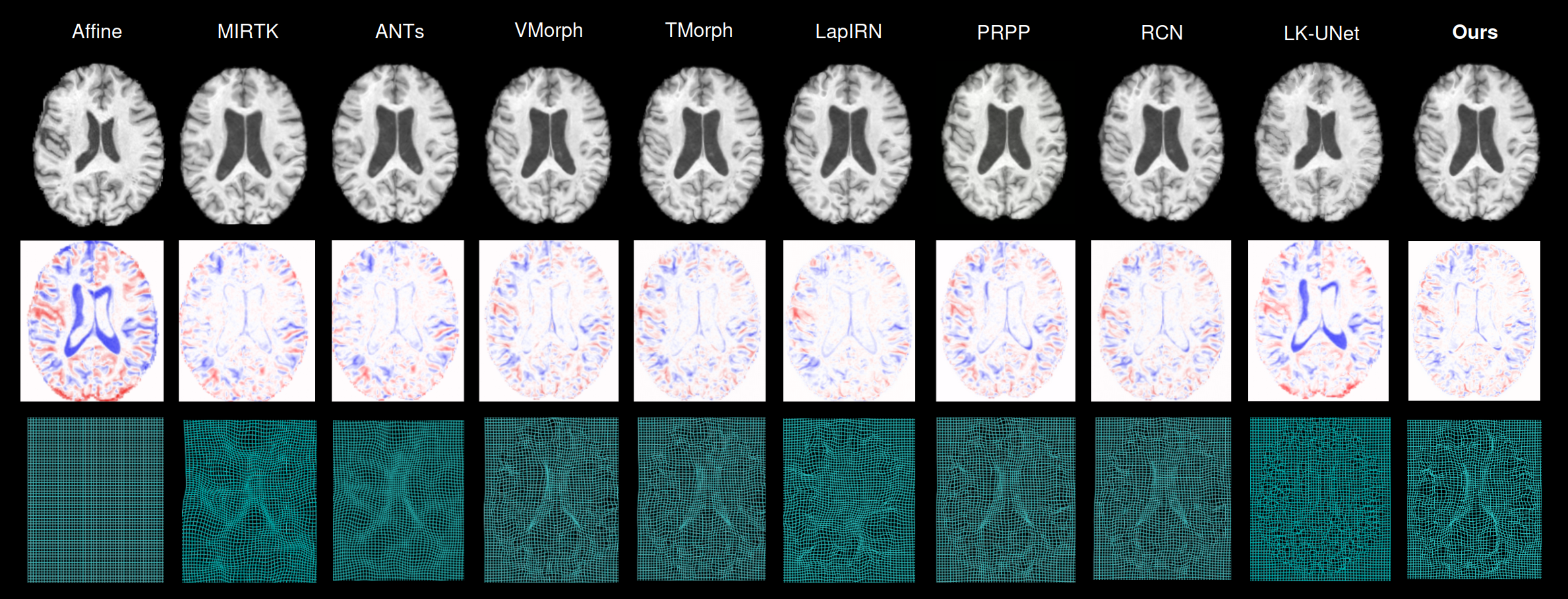}
  \caption{Qualitative results of all compared methods for the CamCAN T1w-T1w inter-subject deformable registration experiment.}
  \label{App:fig:b_qual1}

\end{figure} 

\begin{figure}[b]
  \centering
  \includegraphics[width=1\textwidth]{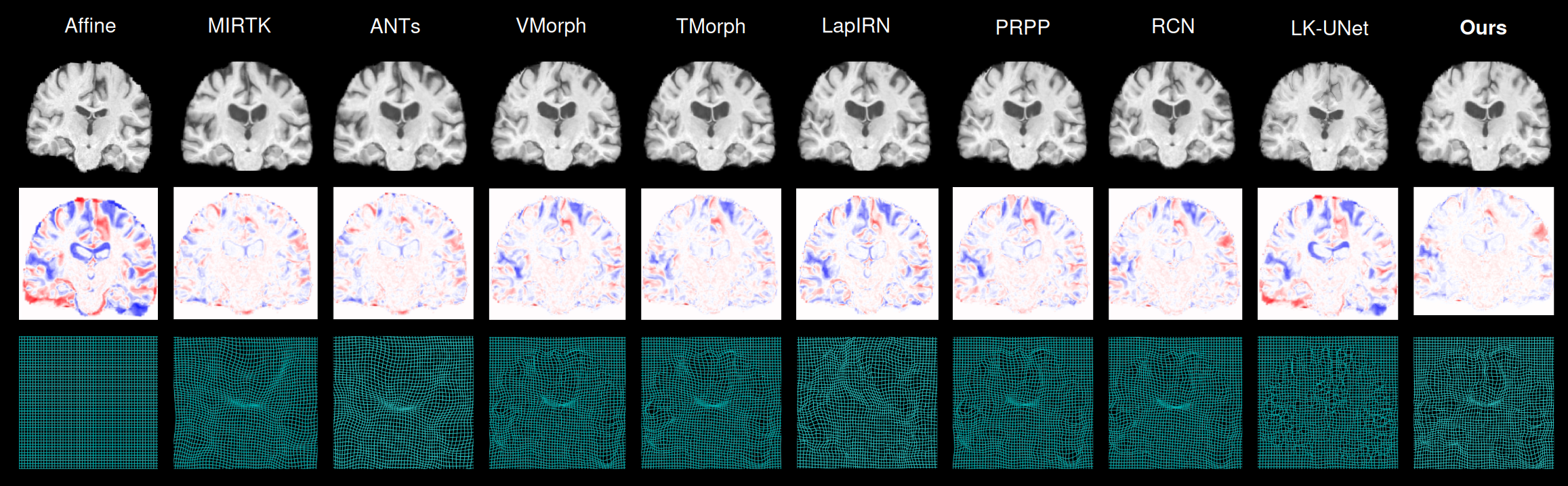}
  \caption{Qualitative results of all compared methods for the CamCAN T1w-T1w inter-subject deformable registration experiment.}
  \label{App:fig:b_qual2}

\end{figure} 

\begin{figure}[b]
  \centering
  \includegraphics[width=1\textwidth]{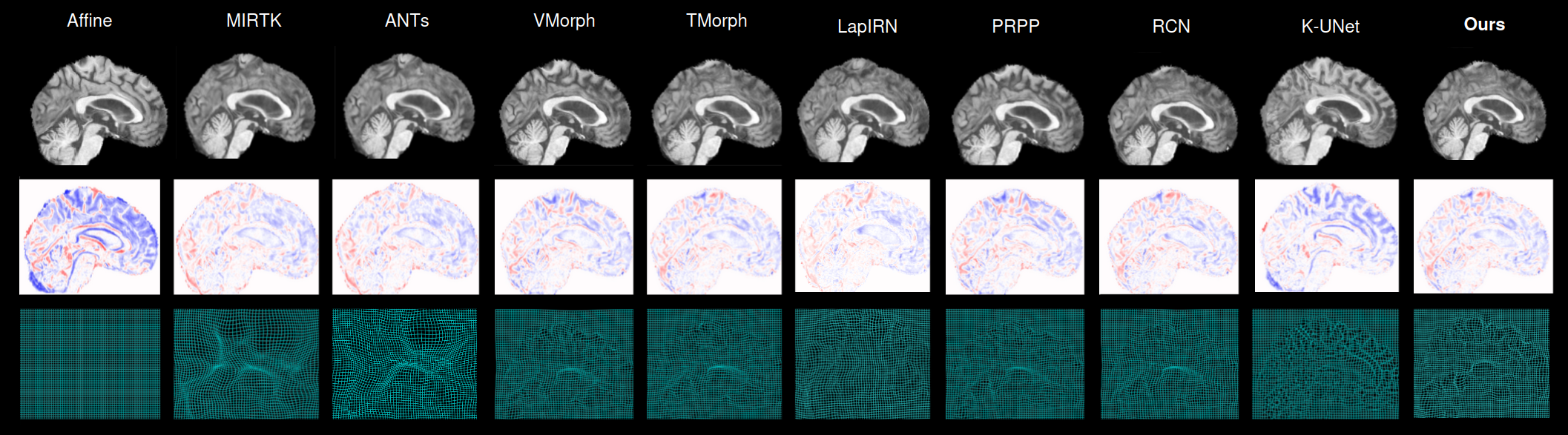}
  \caption{Qualitative results of all compared methods for the CamCAN T1w-T1w inter-subject deformable registration experiment.}
  \label{App:fig:b_qual3}

\end{figure} 

\begin{figure}
  \centering
  \includegraphics[width=1\textwidth]{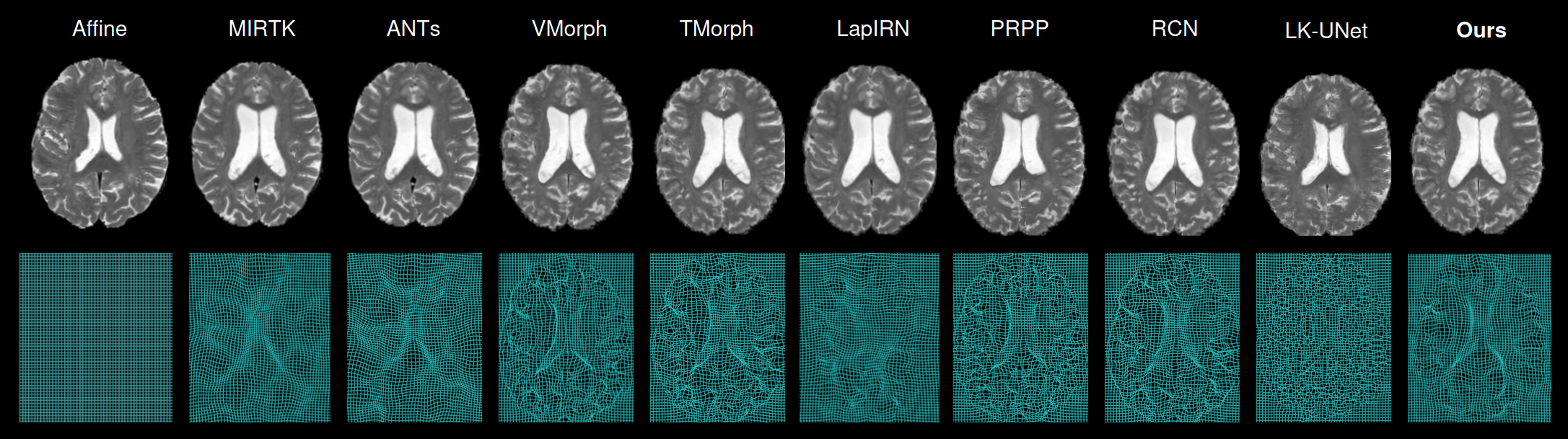}
  \caption{Qualitative results of all compared methods for the CamCAN T1w-T2w inter-subject deformable registration experiment.}
  \label{App:fig:c_qual1}

\end{figure} 

\begin{figure}
  \centering
  \includegraphics[width=1\textwidth]{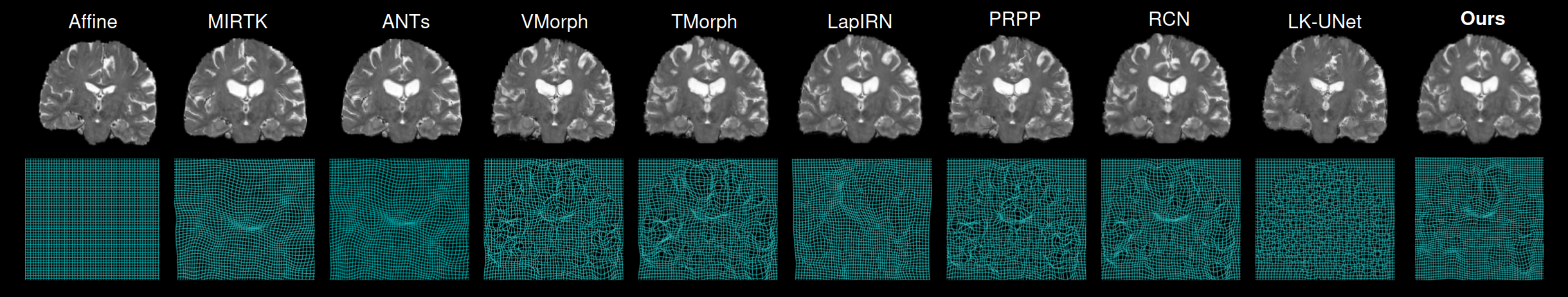}
  \caption{Qualitative results of all compared methods for the CamCAN T1w-T2w inter-subject deformable registration experiment.}
  \label{App:fig:c_qual2}

\end{figure} 

\begin{figure}
  \centering
  \includegraphics[width=1\textwidth]{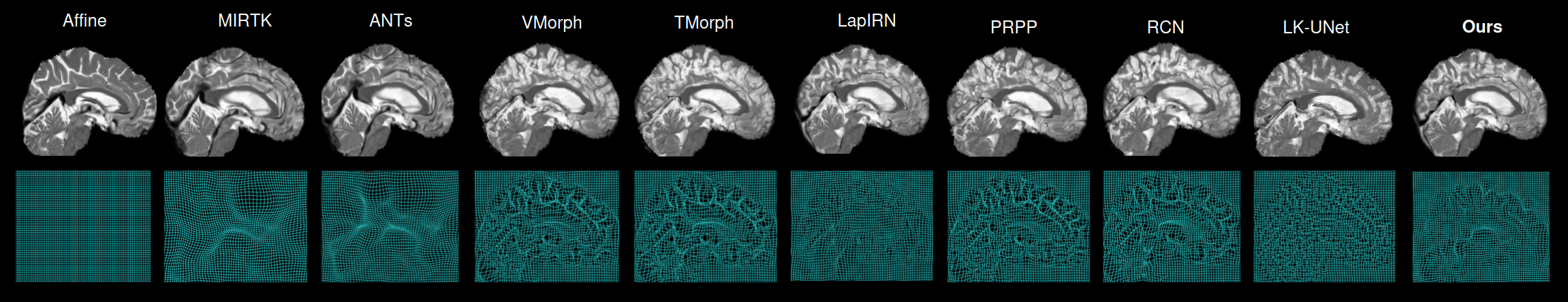}
  \caption{Qualitative results of all compared methods for the CamCAN T1w-T2w inter-subject deformable registration experiment.}
  \label{App:fig:c_qual3}

\end{figure} 

\begin{figure}
  \centering
  \includegraphics[width=1\textwidth]{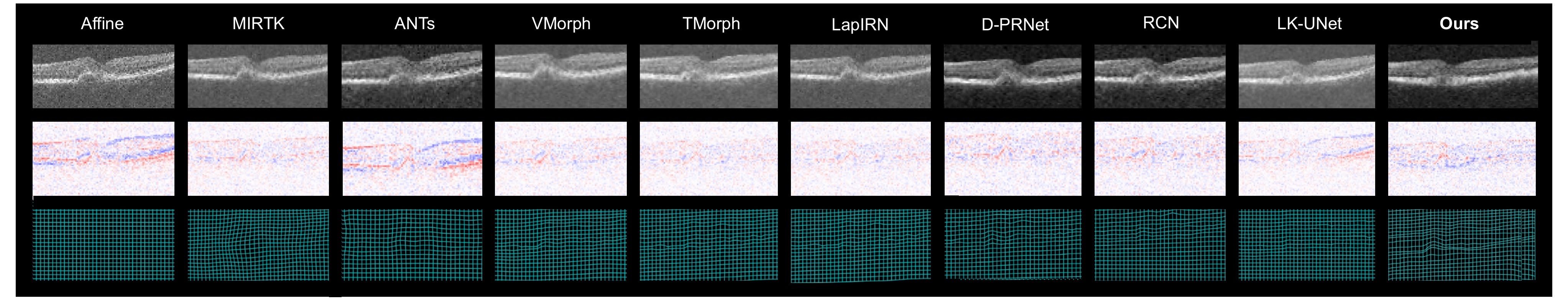}
  \caption{Qualitative results of all compared methods for the retinal optical coherence tomography (OCT) longitudinal deformable registration experiment.}
  \label{App:fig:oct_qual}

\end{figure}

\end{document}